\documentclass{article}

\usepackage{times}
\usepackage{graphicx} 
\usepackage{subfigure} 

\usepackage{natbib}

\usepackage{algorithm}
\usepackage{algorithmic}

\usepackage{hyperref}

\usepackage[accepted]{icml2017}

\usepackage{xspace}
\usepackage{enumitem}
\usepackage{threeparttable}
\usepackage{amsmath}
\usepackage{amssymb}
\usepackage{amsthm}
\usepackage{caption}
\usepackage{epsfig}
\usepackage{bm}
\usepackage{multirow}
\usepackage{verbatim}
\usepackage{soul, xcolor}
\usepackage{color}

\renewcommand{\algorithmicrequire}{\textbf{Input:}}
\renewcommand{\algorithmicensure}{\textbf{Output:}}
\renewcommand\algorithmiccomment[1]{{\small/*{\it#1}*/}}

\icmltitlerunning{Learning Hawkes Processes from Short Doubly-Censored Event Sequences}

\begin{document} 

\twocolumn[
\icmltitle{Learning Hawkes Processes from Short Doubly-Censored Event Sequences}

\icmlauthor{Hongteng Xu}{hxu42@gatech.edu}
\icmladdress{Georgia Institute of Technology}
\icmlauthor{Dixin Luo}{dixin.luo@utoronto.ca}
\icmladdress{University of Toronto}
\icmlauthor{Hongyuan Zha}{zha@cc.gatech.edu}
\icmladdress{Georgia Institute of Technology}

\icmlkeywords{Hawkes process, doubly-censored event sequence, sampling-stitching, data synthesis}

\vskip 0.3in
]

\begin{abstract}
Many real-world applications require robust algorithms to learn point processes based on a type of incomplete data --- the so-called short doubly-censored (SDC) event sequences. 
We study this critical problem of quantitative asynchronous event sequence analysis under the framework of Hawkes processes by leveraging the idea of data synthesis. 
Given SDC event sequences observed in a variety of time intervals, we propose a sampling-stitching data synthesis method --- sampling predecessors and successors for each SDC event sequence from potential candidates and stitching them together to synthesize long training sequences. 
The rationality and the feasibility of our method are discussed in terms of arguments based on likelihood. 
Experiments on both synthetic and real-world data demonstrate that the proposed data synthesis method improves learning results indeed for both time-invariant and time-varying Hawkes processes.
\end{abstract} 

\section{Introduction}
\label{sec:intro}
Real-world interactions among multiple entities are often recorded as asynchronous event sequences, such as user behaviors in social networks, job hunting and hopping among companies, and diseases and their complications. 
The entities or event types in the sequences often exhibit self-triggering and mutually-triggering patterns. 
For example, a tweet of a twitter user may trigger further responses from her friends~\cite{zhao2015seismic}. 
A disease of a patient may trigger other complications~\cite{choi2015constructing}. 
Hawkes processes, an important kind of temporal point process model~\cite{hawkes1974cluster}, have capability to describe the triggering patterns quantitatively and capture the infectivity network of the entities.

Despite the usefulness of Hawkes processes, robust learning of Hawkes processes often needs many event sequences with events occurring over a long observation window. 
Unfortunately, the observation window is likely to be very short and sequence-specific in many important practical applications, i.e., within an imagined universal window, each sequence is only observed with a 
corresponding short sub-interval of it,
and the events outside this sub-interval are not observed --- we call them 
short doubly-censored (SDC) event sequences. 
Existing learning algorithms of Hawkes processes directly applied to SDCs may suffer from over-fitting, and  
what is worse, the triggering patterns between historical events and current ones are lost, so that the triggering patterns learned from SDC event sequences are often unreliable. 
This problem is a thorny issue in several practical applications, especially in those having time-varying triggering patterns. 
For example, the disease networks of patients should evolve with the increase of age. 
However, it is very hard to track and record people's diseases on a life-time scale.
Instead, we can only obtain their several admissions (even only one admission) in a hospital during one or two years, which are just SDC event sequences. 
Therefore, it is highly desirable to propose a method to learn Hawkes processes having a longtime support from a collection of SDC event sequences

\begin{figure*}[t]
\centering
\includegraphics[width=0.13\linewidth]{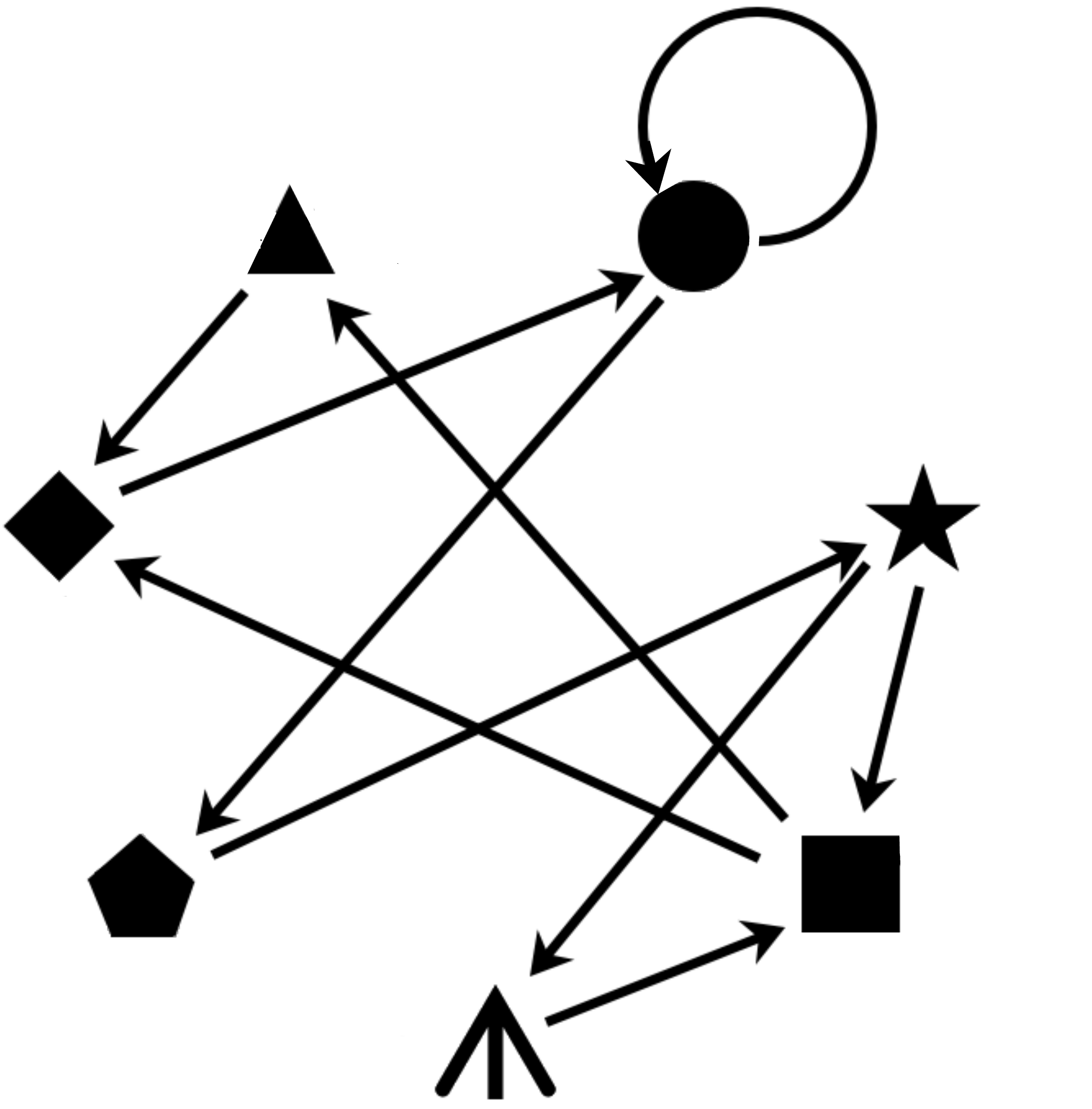}~
\includegraphics[width=0.65\linewidth]{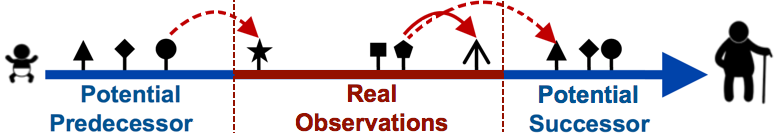}~
\includegraphics[width=0.13\linewidth]{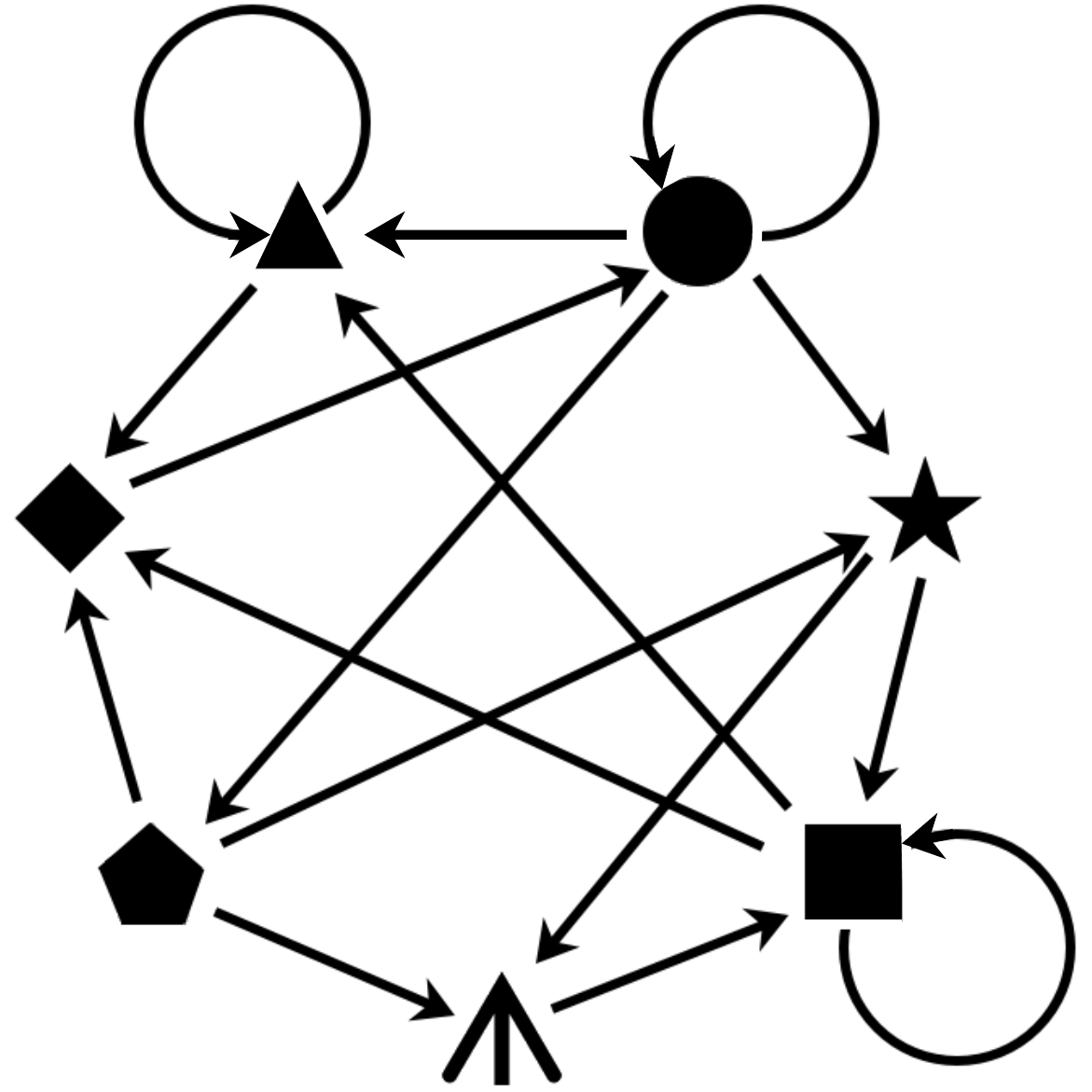}
\vspace{-7pt}
\caption{An illustration of our sampling-stitching data synthesis method. 
For each SDC sequence, i.e., incomplete disease history of a person in his lifetime, we design a mechanism to select other SDC sequences as predecessors/successors and synthesize a long sequence. 
Then, we can estimate the unobserved triggering patterns among diseases, i.e., the red dashed arrows, and construct a dynamical disease network changing over age.}
\label{illu0}
\end{figure*}

In this paper, we propose a novel and simple data synthesis method to enhance the robustness of learning algorithms for Hawkes processes. 
Fig.~\ref{illu0} illustrates the principle of our method.
Given a set of SDC event sequences, we sample predecessor for each event sequence from potential candidates and stitch them together as new training data. 
In the sampling step, the distribution of predecessor (and successor) is estimated according to the similarities between current sequence and its candidates, and the similarity is defined based on the information of time stamps and (optional) features of event sequences. 
We analyze the rationality and the feasibility of our data synthesis method and discuss the necessary condition for using the method. 
Experimental results show that our data synthesis method indeed helps to improve the robustness of various learning algorithms for Hawkes processes. 
Especially in the case of time-varying Hawkes processes, applying our method in the learning phase achieves much better results than learning directly from SDC event sequences, which is meaningful for many practical applications, e.g., constructing dynamical disease network, and learning long-term infectivity among different IT companies.

\section{Related Work}\label{sec:relate}
An event sequence can be represented as $\bm{s}=\{(t_i,c_i)\}^{M}_{i=1}$, where time stamps $t_i$'s are in an observation window $[T_b,T_e]$ and events $c_i$'s are in a set of event types $\mathcal{C}=\{1,...,C\}$. 
A point process $\{N_c\}_{c\in \mathcal{C}}$ is a random process model taking event sequences as instances, where $N_c=\{N_c(t)|t\in [T_b, T_e]\}$ and $N_c(t)$ is the number of type-$c$ events occurring at or before time $t$. 
A point process can be characterized via its conditional intensity function  
$\lambda_{c}(t) = \mathbb{E}[dN_c(t)|\mathcal{H}_t^{\mathcal{C}}]/dt$, where  $c\in\mathcal{C}$ and $\mathcal{H}_t^{\mathcal{C}}=\{(t_i, c_i)|t_i<t, c_i\in\mathcal{C}\}$ is the set of history. 
It represents the expected instantaneous happening rate of events given historical record~\cite{daley2007introduction}. 
The intensity is often modeled with certain parameters $\bm{\Theta}$ to capture the phenomena of interests, i.e., self-triggering~\cite{hawkes1974cluster} or self-correcting~\cite{xu2015trailer}. 
Based on $\{\lambda_c(t)\}_{c\in\mathcal{C}}$, the likelihood of an event sequence $\bm{s}$ is 
\begin{eqnarray}\label{like}
\begin{aligned}
\mathcal{L}(\bm{s};\bm{\Theta})=\sideset{}{_{i}}\prod\lambda_{c_i}(t_i)\exp\Bigl(-\sideset{}{_{c}}\sum\int_{T_b}^{T_e}\lambda_c(s)ds\Bigr).
\end{aligned}
\end{eqnarray}

\textbf{Hawkes Processes.} Hawkes processes~\cite{hawkes1974cluster} have a particular form of intensity:
\begin{eqnarray}\label{intensity}
\begin{aligned}
\lambda_{c}(t) = \mu_c + \sideset{}{_{c'=1}^{C}}\sum\int_{0}^{t}\phi_{cc'}(t,s)d N_{c'}(s),
\end{aligned}
\end{eqnarray}
where $\mu_c$ is the exogenous base intensity independent of the history while $\int_{0}^{t}\phi_{cc'}(t,s)d N_{c'}(s)$ is the endogenous intensity capturing the influence of historical events on type-$c$ ones at time $t$~\cite{xu2016learning}. 
Here, $\phi_{cc'}(t,s)\geq 0$ is called \emph{impact function}. 
It quantifies the influence of the type-$c'$ event at time $s$ to the type-$c$ event at time $t$. 
Hawkes processes provide us with a physically-meaningful model to capture the infectivity among various events, which are used in social network analysis~\cite{zhou2013learning2,zhao2015seismic}, behavior analysis~\cite{yang2013mixture,luo2015multi} and financial analysis~\cite{bacry2013some}. 
However, the methods in these references assume that the impact function is shift-invariant (i.e., $\phi_{cc'}(t,s)=\phi_{cc'}(t-s)$, $t\geq s$), which limits their applications on longtime scale. 
Recently, the time-dependent Hawkes process (TiDeH) in~\cite{kobayashi2016tideh} and the neural network-based Hawkes process in~\cite{mei2016neural} learn very flexible Hawkes processes with complicated intensity functions. 
Because they highly depend on the size and the quality of data, they may fail in the case of SDC event sequences.

\textbf{Learning from Imperfect Observations.} In practice, we need to learn sequential models from imperfect observations (e.g., interleaved~\cite{xu2016pinfer}, aggregated~\cite{luo2016learning} and extremely-short sequences~\cite{xu2016icu}). 
Multiple imputation (MI)~\cite{rubin2009multiple} is a general framework to build surrogate observations from the current model. 
For time series, bootstrap method~\cite{efron1982jackknife,politis1994stationary,gonccalves2004bootstrapping} and its variants~\cite{paparoditis2001tapered,guan2007thinned} have been used to improve learning results when observations are insufficient. 
In survival analysis, many techniques have been made to deal with truncated and censored data~\cite{turnbull1974nonparametric,de1989analysis,klein2005survival,van2016inference}. 
For point processes, the global~\cite{streit2010poisson} or local~\cite{fan2009local} likelihood maximization estimators (MLE) are used to learn Poisson processes.
Nonparametric approaches for non-homogeneous Poisson processes use the pseudo MLE~\cite{sun1995estimation} or full MLE~\cite{wellner2000two}.
The bootstrap methods above are also used to learn point processes~\cite{cowling1996bootstrap,guan2007thinned,kirk2009gaussian}. 
To learn Hawkes processes robustly, structural constraints, e.g., low-rank~\cite{luo2015multi} and group-sparse regularizers~\cite{xu2016learning}, are introduced. 
However, all of these methods do not consider the case of SDC event sequences for Hawkes processes. 

\section{Learning from SDC Event Sequences}
\label{sec:learning}
Suppose that the original complete event sequences are in a long observation window. 
However, the observation window in practice might be segmented into several intervals $\{T_b^n, T_e^n\}_{n=1}^{N}$, and we can only observe $K_n$ SDC sequences $\{\bm{s}_k^n\}_{k=1}^{K_n}$ in the $n$-th interval, $n=1,...,N$. 
Although we can still apply maximum likelihood estimator to learn Hawkes processes, i.e., 
\begin{eqnarray}\label{MLEd}
\begin{aligned}
\sideset{}{_{\bm{\Theta}}}\min-\sideset{}{_{n,k}}\sum\log\mathcal{L}(\bm{s}_k^n;\bm{\Theta}), 
\end{aligned}
\end{eqnarray}
the SDC event sequences would lead to over-fitting problem and the loss of triggering patterns. 
Can we do better in such a situation?
In this work, we propose a data synthesis method based on a sampling-stitching mechanism, which extends SDC event sequences to longer ones and enhances the robustness of learning algorithms.

\subsection{Data Synthesis via Sampling-Stitching}\label{ssec:ss}
Denote the $k$-th SDC event sequence in the $n$-th interval as $\bm{s}_k^n$. 
Because its \emph{predecessor} is unavailable, if we learn the parameters of our model via (\ref{MLEd}) directly, we actually impose a strong assumption on our data that there is no event happening before $\bm{s}_k^n$ (or previous events are too far away from $\bm{s}_k^n$ to have influences on $\bm{s}_k^n$). 
Obviously, this assumption is questionable --- it is likely that there are influential events happening before $\bm{s}_k^n$. 
A more reasonable strategy is enumerating potential predecessors and maximizing the expected log-likelihood over the whole observation window: 
\begin{eqnarray}\label{EMLE}
\begin{aligned}
\sideset{}{_{\bm{\Theta}}}\min-\sideset{}{_{n,k}}\sum\mathbb{E}_{\bm{s}\sim \mathcal{H}_{T_b^n}^{\mathcal{C}}}[\log\mathcal{L}([\bm{s},\bm{s}_k^n];\bm{\Theta})].
\end{aligned}
\end{eqnarray}
Here $\mathbb{E}_{x\sim\mathcal{D}}[f(x)]$ represents the expectation of function $f(x)$ with random variable $x$ yielding to a distribution $\mathcal{D}$. 
$\bm{s}\sim\mathcal{H}_{T_b^n}^{\mathcal{C}}$ means all possible history before $T_b^n$, and $\mathcal{L}([\bm{s},\bm{s}_k^n];\bm{\Theta})$ is the likelihood of stitched sequence $[\bm{s},\bm{s}_k^n]$.

The stitched sequence $[\bm{s},\bm{s}_k^n]$ can be generated via \textbf{sampling} SDC sequence $\bm{s}$ from previous $1$st, ..., $(k-1)$-th intervals and \textbf{stitching} $\bm{s}$ to $\bm{s}_k^n$. 
The sampling process yields to the probabilistic distribution of the stitched sequences. 
Given $\bm{s}_k^n$, we can compute its similarity between its potential predecessor $\bm{s}_{k'}^{n'}$ in $[T_b^{n'}, T_e^{n'}]$ as
\begin{eqnarray}\label{sim}
\begin{aligned}
w(\bm{s}_{k'}^{n'},\bm{s}_k^n)=
\begin{cases}
S(T_b^n,T_e^{n'})\underbrace{S(f_k^n,f_{k'}^{n'})}_{\text{\tiny{optional}}}, & T_e^{n'}\leq T_b^{n},\\
0, & \text{otherwise}.
\end{cases}
\end{aligned}
\end{eqnarray} 
Here, $S(a,b)=\exp(-\|b-a\|_2^2/\sigma_s)$ is a predefined similarity function with parameter $\sigma_s$. 
$f_k^n$ is the feature of $\bm{s}_k^n$, which is available in some applications. 
Note that the availability of feature is optional --- even if the feature of sequence is unavailable, we can still define the similarity measurement purely based on time stamps.
The normalized $\{w(\bm{s}_{k'}^{n'},\bm{s}_k^n)\}$ provides us with the probability that $\bm{s}_{k'}^{n'}$ appears before $\bm{s}_k^n$, i.e., $p(\bm{s}_{k'}^{n'}|\bm{s}_k^n)\propto w(\bm{s}_{k'}^{n'},\bm{s}_k^n)$.
Then, we can sample $\bm{s}_{k'}^{n'}$ according to the categorical distribution, i.e., $\bm{s}_{k'}^{n'}\sim\mbox{Category}(w(\cdot,\bm{s}_k^n))$.


We can apply such a sampling-stitching mechanism $L$ times iteratively to the SDC sequences in both backward and forward directions and get long stitched event sequences. 
Specifically, we represent a stitched event sequence as $\bm{s}_{stitch}=[\bm{s}_1,...,\bm{s}_{2L+1}]$, $\bm{s}_l\in\{\bm{s}_{k}^n\}$, $l=1,...,2L+1$, whose probability is
\begin{eqnarray}\label{prob}
\begin{aligned}
p(\bm{s}_{stitch})\propto \sideset{}{_{l=1}^{2L}}\prod w(\bm{s}_l,\bm{s}_{l+1}).
\end{aligned}
\end{eqnarray}

Note that our data synthesis method naturally contains two variants. 
When the starting (the ending) point of time window is unavailable, we use the time stamp of the first (the last) event of SDC sequence instead. 
Additionally, we can relax the constraint $T_e^{n'}\leq T_b^{n}$ in~(\ref{sim}) and allow a SDC sequence to have an overlap with its predecessor/successor. 
In this case, we preserve the overlap part randomly either from itself or its predecessor/successor before applying our sampling-stitching method. 
These two variants ensure that our data synthesis method is doable in practice, which are used in the following experiments on real-world data.

\subsection{Justification}
After applying our data synthesis method, we obtain many stitched event sequences, which can be used as instances for estimating $\mathbb{E}_{\bm{s}\sim \mathcal{H}_{T_b^n}^{\mathcal{C}}}[\log\mathcal{L}([\bm{s},\bm{s}_k^n];\bm{\Theta})]$. 
Specifically, taking advantage of stitched sequences, we can rewrite the learning problem in~(\ref{EMLE}) approximately as
\begin{eqnarray}\label{cross}
\begin{aligned}
\sideset{}{_{\bm{\Theta}}}\min-\sideset{}{_{\bm{s}_{stitch}\in\bm{S}}}\sum p(\bm{s}_{stitch})\log\mathcal{L}(\bm{s}_{stitch};\bm{\Theta}),
\end{aligned}
\end{eqnarray}
which is actually the minimum cross-entropy estimation. 
$p(\bm{s}_{stitch})$ represents the ``true'' probability that the stitched sequence happens, which is estimated via the predefined similarity measurement and the sampling mechanism. 
The likelihood $\mathcal{L}(\bm{s}_{stitch};\bm{\Theta})$ represents the ``unnatural'' probability that the stitched sequence happens, which is estimated based on the definition in~(\ref{like}). 
Our data synthesis method takes advantage of the information of time stamps and (optional) features and makes $p(\bm{s}_{stitch})$ suitable for practical situations. 
For example, the likelihood of a sequence generally reduces with the increase of observation time window. 
The proposed probability $p(\bm{s}_{stitch})$ yields to the same pattern --- according to~(\ref{prob}), the longer a stitched sequence is, the smaller its probability becomes.

The set of all possible stitched sequences, i.e., the $\bm{S}$ in~(\ref{cross}), is very large, whose cardinality is $|\bm{S}|=\mathcal{O}(\prod_{n=1}^{N}K_n)$. 
In practice, we cannot and do not need to enumerate all possible combinations. 
An empirical setting is making the number of stitched sequences comparable to that of original SDC event sequences, i.e., generating $\mathcal{O}(\sum_{n=1}^{N}K_n)$ stitched sequences.
In the following experiments, we just apply $5$($=U$) trials and generate $5$ stitched sequences for each original SDC event sequence, which achieves a trade-off between computational complexity and performance.

\subsection{Feasibility}
It should be noted that our data synthesis method is only suitable for those complicated point processes whose historical events have influences on current and future ones. 
Specifically, we analyze the feasibility of our method for several typical point processes.

\textbf{Poisson Processes.} Our data synthesis method cannot improve learning results if the event sequences are generated via Poisson processes. 
For Poisson processes, the happening rate of future events is independent of historical events. 
In other words, the intensity function of each interval can be learned independently based on the SDC event sequences. 
The stitched sequences do not provide us with any additional information. 

\textbf{Hawkes Processes.} For Hawkes processes, whose intensity function is defined as~(\ref{intensity}), our data synthesis method can enhance the robustness of learning algorithm generally. 
In particular, consider a ``long'' event sequence generated via a Hawkes process in the time window $[T_b, T_e]$. 
If we divide the time window into $2$ intervals, i.e., $[T_b,T]$ and $(T,T_e]$, the intensity function corresponding to the second interval can be written as
\begin{eqnarray}\label{int2}
\begin{aligned}
\lambda_{c}(t) =\mu_c + \sum_{t_i\leq T}\phi_{cc_i}(t, t_i)
+\sum_{T<t_i\leq T_e}\phi_{cc_i}(t,t_i).
\end{aligned}
\end{eqnarray}
If the events in the first interval are unobserved, we just have a SDC event sequence, and the second term in~(\ref{int2}) is unavailable. 
Learning Hawkes processes directly from the SDC event sequence ignores the information of the second term, which has a negative influence on learning results. 
Our data synthesis method leverages the information from other potential predecessors and generates multiple candidate long sequences. 
As a result, we obtain multiple intensity functions sharing the second interval and maximize the weighted sum of their log-likelihood functions (i.e., an estimated expectation of the log-likelihood of the real long sequence), as~(\ref{cross}) does. 

Compared with learning from SDC event sequences directly, applying our data synthesis method can improve learning results in general, unless the term $\sum_{t_i\leq T}\phi_{cc_i}(t, t_i)$ is ignorable. 
Specifically, we can model the impact functions $\{\phi_{cc'}(t,s)\}_{c,c'\in\mathcal{C}}$ of Hawkes processes based on basis representation:
\begin{eqnarray}\label{tvimpact}
\begin{aligned}
\phi_{cc'}(t,s)&=
\underbrace{\psi_{cc'}(t)}_{
\text{Infectivity}}\times
\underbrace{g(t-s)}_{
\text{Triggering kernel}}\\
&=\sideset{}{_{m=1}^{M}}\sum a_{cc'm}\kappa_{m}(t)g(t-s).
\end{aligned}
\end{eqnarray}
Here, we decompose impact functions into two parts: 
1) \emph{Infectivity} $\psi_{cc'}(t)=\sum_{m=1}^{M}a_{cc'm}\kappa_m(t)$ represents the infectivity of event type $c'$ to $c$ at time $t$.\footnote{When $M=1$ and $\kappa_m(t)\equiv 1$, we obtain the simplest time-invariant Hawkes process. 
Relaxing the shift-invariant assumption, i.e., $M>1$ and $\kappa_m(t)$ is Gaussian, we obtain a flexible time-varying Hawkes process model.}
2) \emph{Triggering kernel} $g(t)=\exp(-\beta t)$ measures the time decay of infectivity. 
It means that the infectivity of a historical event to current one reduces exponentially with the increase of temporal distance between them. 
When $\beta$ is very large, $\phi_{cc'}(t,s)$ decays rapidly with the increase of $t-s$, and the events happening long ago can be ignored. 
In such a situation, our data synthesis method is unable to improve learning results.

\section{Implementation for Hawkes Processes}
Hawkes process is a kind of physically-interpretable model for many natural and social phenomena. 
The proposed model in~(\ref{tvimpact}) reflects many common properties of real-world event sequences. 
First, the infectivity among various event types often changes smoothly in practice: in social networks, the interaction between two users changes smoothly, which is not established or blocked suddenly; in disease networks, the infectivity among diseases should change smoothly with the increase of patient's age. 
Applying Gaussian basis representation guarantees the smoothness of infectivity function. 
Second, the triggering kernel measures the decay of infectivity over time. 
According to existing work, the decay of infectivity is exponential approximately, which has been verified in many real-world data~\cite{zhou2013learning,kobayashi2016tideh,choi2015constructing}. 
For learning Hawkes processes from SDC event sequences, we combine our data synthesis method with an EM-based learning algorithm of Hawkes processes.
Applying our data synthesis method, we obtain a set of stitched event sequences $\bm{S}=\{\bm{s}_n\}$ and their appearance probabilities $\{p_n\}$, where $\bm{s}_n=\{(t_i^n, c_i^n)_{i=1}^{I_n}| t_i^n\in [T_b^n, T_e^n],~c_i^n\in\mathcal{C}\}$ and $p_n$ is calculated based on~(\ref{sim}). 
According to (\ref{cross},~\ref{tvimpact}), we can learn target Hawkes process via 
\begin{eqnarray}\label{MLE}
\begin{aligned}
\min_{\bm{\mu}\geq\bm{0},~\bm{A}\geq\bm{0}}~-\sideset{}{_{n=1}^{|\bm{S}|}}\sum p_n\log\mathcal{L}(\bm{s}_n;\bm{\Theta})+\gamma\mathcal{R}(\bm{A}).
\end{aligned}
\end{eqnarray}
Here $\bm{\Theta}=\{\bm{\mu},~\bm{A}\}$ represents the parameters of our model.
The vector $\bm{\mu}=[\mu_c]$ and the tensor $\bm{A}=[a_{cc'm}]$ are nonnegative. 
Based on~(\ref{like},~\ref{tvimpact}), the log-likelihood function is
\begin{eqnarray*}\label{loglike}
\begin{aligned}
&\log\mathcal{L}(\bm{s}_n;\bm{\Theta})\\
=&\sideset{}{_{i=1}^{I_n}}\sum\log\lambda_{c_i}(t_i) -\sideset{}{_{c=1}^{C}}\sum\int_{T_b^n}^{T_e^n}\lambda_c(s)ds\\
=&\sideset{}{_{i=1}^{I_n}}\sum\log\biggl[\mu_{c_i^n}
+\sideset{}{_{j<i}}\sum g(\tau_{ij}^n)\sideset{}{_{m=1}^{M}}\sum a_{c_i^nc_j^nm}
\kappa_m(t_i^n)\biggr]\\
&-\sideset{}{_{c=1}^{C}}\sum\Bigl(\Delta^n\mu_c-\sideset{}{_{m=1}^{M}}\sum
\sideset{}{_{i=1}^{I_n}}\sum\sideset{}{_{j\leq i}}\sum a_{cc_j^nm}
G_{ij}\Bigr),
\end{aligned}
\end{eqnarray*}
where $\tau_{ij}^n=t_i^n-t_j^n$, $G_{ij}=\int_{t_i^n}^{t_{i+1}^n}\kappa_m(s)g(s-t_j^n)ds$, and $\Delta^n = T_{e}^n-T_{b}^n$. 
$\mathcal{R}(\bm{A})$ represents the regularizer of parameters, whose weight is $\gamma$. 
Following existing work in~\cite{luo2015multi,zhou2013learning,xu2016learning}, we assume the infectivity connections among different event types to be sparse and impose a $\ell_1$-norm regularizer on the coefficient tensor $\bm{A}$, i.e., $\mathcal{R}(\bm{A})=\|\bm{A}\|_1=\sum_{c,c',m}|a_{cc'm}|$. 
 
We can solve the problem via an EM algorithm. 
Specifically, when sparse regularizer is applied, we take advantage of ADMM method, introducing auxiliary variable $\bm{Z}=[z_{cc'm}]$ and dual variable $\bm{U}=[u_{cc'm}]$ for $\bm{A}$ and rewriting the objective function in (\ref{MLE}) as
\begin{eqnarray*}\label{opt}
\begin{aligned}
&-\sideset{}{_n}\sum p_n\log\mathcal{L}(\bm{s}_n;\bm{\Theta})+0.5\rho\|\bm{A}-\bm{Z}\|_F^2\\
&\quad\quad+\rho\mbox{tr}(\bm{U}^{\top}(\bm{A}-\bm{Z}))+\gamma\|\bm{Z}\|_1.
\end{aligned}
\end{eqnarray*} 
Here $\rho$ controls the weights of regularization terms, which increases with the number of EM iterations. 
$\mbox{tr}(\cdot)$ computes the trace of matrix.
Then, we can update $\{\bm{\mu},\bm{A}\}$, $\bm{Z}$, and $\bm{U}$ alternatively. 

\textbf{Update $\bm{\mu}$ and $\bm{A}$:} 
Given the parameters in the $k$-th iteration, we apply Jensen's inequality to $-\sum_n\log\mathcal{L}(\bm{s}_n;\bm{\Theta})$ and obtain a surrogate objective function for $\bm{\mu}$ and $\bm{A}$:
\begin{eqnarray*}
\begin{aligned}
&\mathcal{Q}(\bm{\mu},\bm{A};~\bm{\mu}^k,\bm{A}^k,\bm{Z}^k,\bm{U}^k)\\
=&-\sum_{n=1}^{N}p_n\biggl\{\sum_{i=1}^{I_n}\biggl[
\sum_{j<i}\sum_{m=1}^{M}q_{ijm}\log\frac{g(\tau_{ij}^n)a_{c_i^nc_j^nm}
\kappa_m(t_i^n)}{q_{ijm}}\\
&+q_i\log\frac{\mu_{c_i^n}}{q_i}-\sum_{c=1}^{C}\sum_{m=1}^{M}
\sum_{j\leq i}a_{cc_j^nm}
G_{ij}\biggr]-\Delta^n\sum_{c=1}^{C}\mu_c\biggr\}\\
&+0.5\rho\|\bm{A}-\bm{Z}^k+\bm{U}^k\|_F^2,
\end{aligned}
\end{eqnarray*}
where $q_i=\frac{\mu_{c_i^n}^k}{\lambda_{c_i^n}^k(t_i^n)}$ and $q_{ijm}=\frac{g(\tau_{ij}^n)a_{c_i^nc_j^nm}^{k}
\kappa_m(t_i^n)}{\lambda_{c_i^n}^k(t_i^n)}$, and $\lambda_{c_i^n}^k(t_i^n)$ is calculated based on $\bm{\mu}^k$ and $\bm{A}^k$. 
Then, we can update $\bm{\mu}$ and $\bm{A}$ via solving $\frac{\partial\mathcal{Q}}{\partial \bm{\mu}}=\bm{0}$ and $\frac{\partial\mathcal{Q}}{\partial \bm{A}}=\bm{0}$. 
Both of these two equations have closed-form solution:
\begin{eqnarray}\label{UpdateAmu}
\begin{aligned}
\mu_c^{k+1}=\frac{\sum_{n}p_n\sum_{c_i^n = c}q_i}{\sum_{n}p_n\Delta_n},
a_{cc'm}^{k+1}=\frac{\sqrt{B^2-4\rho C}-B}{2\rho},
\end{aligned}
\end{eqnarray}
where 
\begin{eqnarray*}
\begin{aligned}
B&=\rho(u_{cc'm}^k - z_{cc'm}^k)+\sideset{}{_{n}}\sum p_n\sideset{}{_{c_i^n=c,~c_j^n=c',~j\leq i}}\sum G_{ij},\\
C&=-\sideset{}{_{n}}\sum p_n\sideset{}{_{c_i^n=c,~c_j^n=c',~j\leq i}}\sum q_{ijm}.
\end{aligned}
\end{eqnarray*}

\textbf{Update $\bm{Z}$:} 
Given $\bm{A}^{k+1}$ and $\bm{U}^k$, we can update $\bm{Z}$ via solving the following optimization problem:
\begin{eqnarray*}
\begin{aligned}
\sideset{}{_{\bm{Z}}}\min~\gamma\|\bm{Z}\|_1+ 0.5\rho\|\bm{A}^{k+1}-\bm{Z}+\bm{U}^k\|_F^2.
\end{aligned}
\end{eqnarray*}
Applying soft-thresholding method, we have
\begin{eqnarray}\label{updateZ}
\begin{aligned}
\bm{Z}^{k+1} = F_{\frac{\gamma}{\rho}}(\bm{A}^{k+1}+\bm{U}^k),
\end{aligned}
\end{eqnarray}
where $F_{\eta}(x)=\mbox{sign}(x)\min\{|x|-\eta,0\}$ is the soft-thresholding function. 

\textbf{Update $\bm{U}$:} 
Given $\bm{A}^{k+1}$ and $\bm{Z}^{k+1}$, we can further update dual variable as
\begin{eqnarray}\label{updateU}
\begin{aligned}
\bm{U}^{k+1} = \bm{U}^{k}+(\bm{A}^{k+1}-\bm{Z}^{k+1}).
\end{aligned}
\end{eqnarray}

In summary, Algorithm~\ref{alg2} shows the scheme of our learning method. 
Note that the algorithm can be applied to SDC event sequences directly via ignoring $p_n$'s.
\begin{algorithm}[t]
   \caption{Learning Algorithm of Hawkes Processes}
   \label{alg2}
\begin{algorithmic}[1]
   \STATE \textbf{Input:} Event sequences $\bm{S}$. The threshold $V$. Predefined parameters $\beta$, $\sigma_{\kappa}$, and $\gamma$.
   \STATE \textbf{Output:} Parameters $\bm{A}$ and $\bm{\mu}$. 
   \STATE Initialize $\bm{A}^k$ and $\bm{\mu}^k$ randomly. $\bm{Z}^k=\bm{A}^k$, $\bm{U}^k=\bm{0}$. $k=0$, $\rho=1$.
   \REPEAT
   \STATE Obtain $\bm{A}^{k+1}$ and $\bm{\mu}^{k+1}$ via (\ref{UpdateAmu}).
   \STATE Obtain $\bm{Z}^{k+1}$ via (\ref{updateZ}).
   \STATE Obtain $\bm{U}^{k+1}$ via (\ref{updateU}).
   \STATE $k=k+1$, $\rho=1.5\rho$.
   \UNTIL{$\|\bm{A}^{k}-\bm{A}^{k-1}\|_F<V$}
   \STATE $\bm{A}=\bm{A}^k$, $\bm{\mu}=\bm{\mu}^k$.
\end{algorithmic}
\end{algorithm}

\section{Experiments}\label{sec:exp} 
\subsection{Implementation Details}
To demonstrate the usefulness of our data synthesis method, we combine it with various learning algorithms of Hawkes processes and learn different models accordingly from SDC event sequences. 
For time-invariant Hawkes processes, we consider two learning algorithms --- our EM-based learning algorithm and the least squares (LS) algorithm in~\cite{eichler2016graphical}. 
For time-varying Hawkes processes, we apply our EM-based learning algorithm. 
In the following experiments, we use Gaussian basis functions:  $\kappa_m(t)=\exp((t-t_m)^2/\sigma_{\kappa})$ with center $t_m$ and bandwidth $\sigma_{\kappa}$. 
The number and the bandwidth of basis can be set according to the basis selection method proposed in~\cite{xu2016learning}. 
Additionally, we set $V=10^{-4}$, $\gamma=1$, and $\sigma_{s}=1$ in our algorithm. 
Given SDC event sequences, we learn Hawkes processes in three ways: 1) learning directly from SDC event sequences; 2) applying the stationary bootstrap method in~\cite{politis1994stationary} to generate more synthetic SDC event sequences and learning from these sequences accordingly; 3) learning from stitched sequences generated via our data synthesis method. 
For real-world data, whose SDC sequences do not have predefined starting and ending time stamps, we applied the variants of our method mentioned in the end of Section~\ref{ssec:ss}.

\begin{figure}[t]
\centering
\subfigure[Our learning algorithm]{
\includegraphics[width=0.95\linewidth]{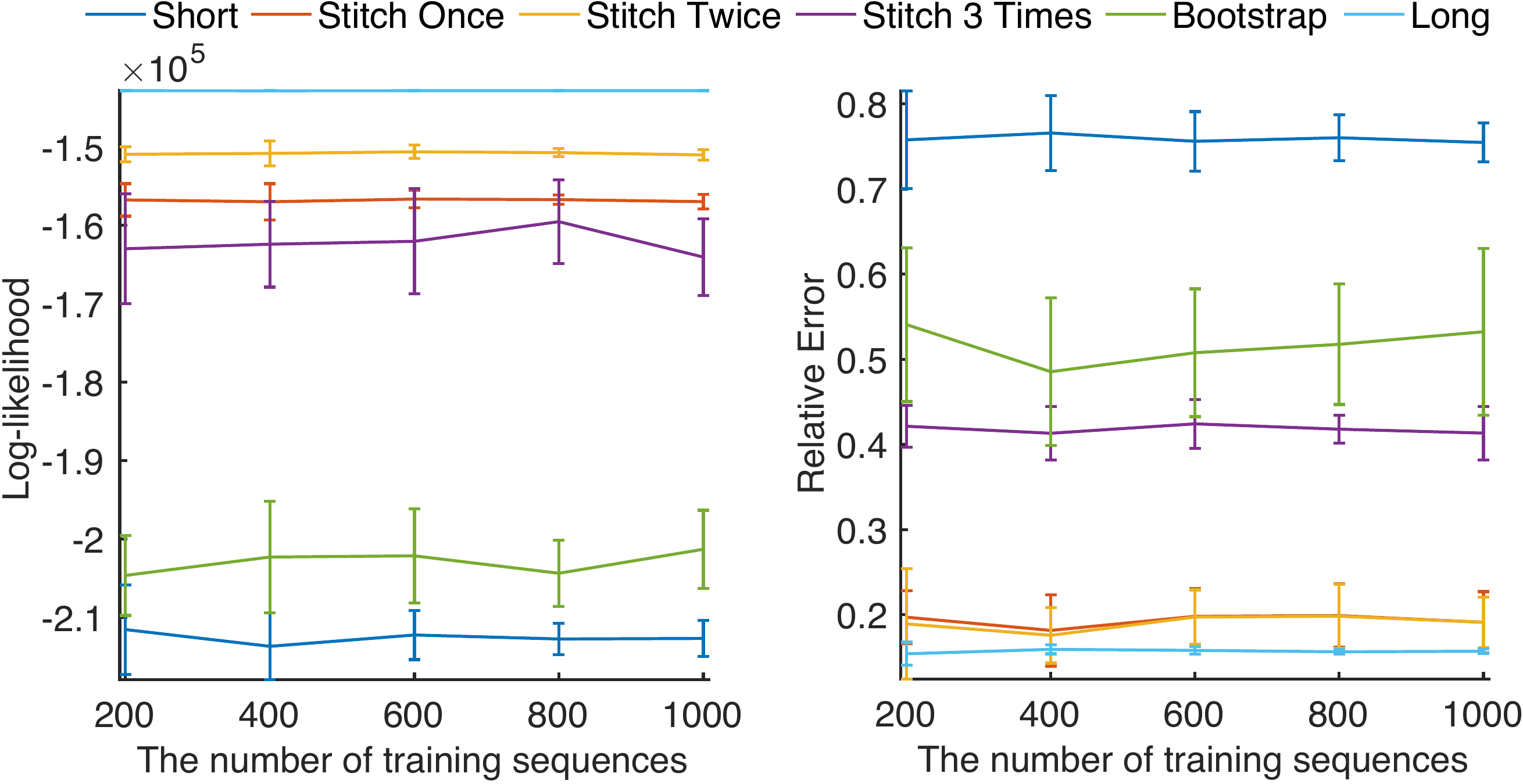}\label{Exp_HP_My}
}
\subfigure[Least squares algorithm]{
\includegraphics[width=0.95\linewidth]{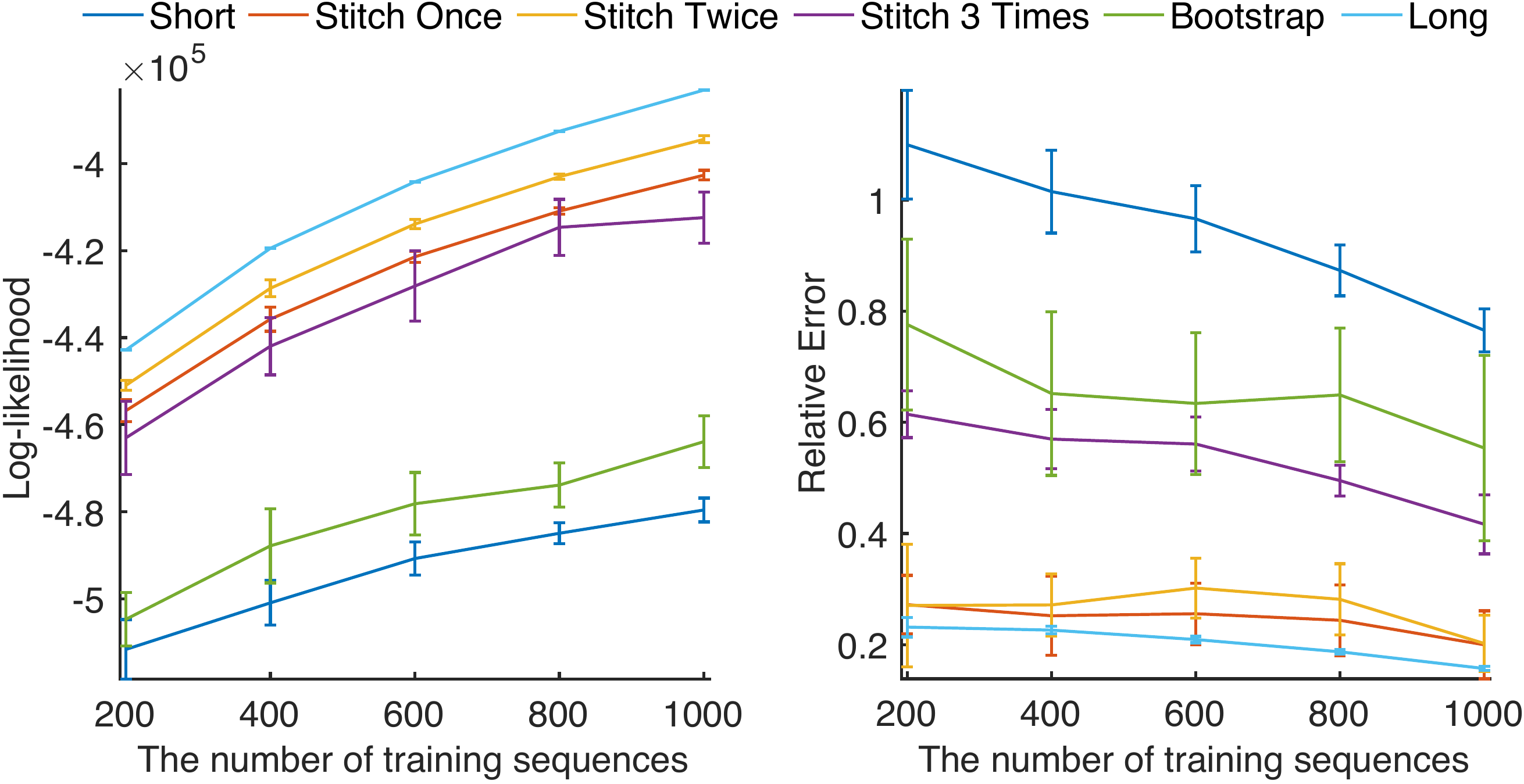}\label{Exp_HP_LS}
}\vspace{-7pt}
\caption{Comparisons on log-likelihood and relative error.}\label{Exp_HP}
\end{figure}

\subsection{Synthetic Data}\label{ssec:syn}
The synthetic SDC event sequences are generated via the following method: $2000$ complete event sequences are simulated in the time window $[0,50]$ based on a $2$-dimensional Hawkes process. 
The base intensity $\{\mu_c\}_{c=1}^{2}$ are randomly generated in the range $[0.1,0.2]$. 
The parameter of triggering kernel, $\beta$, is set to be $0.2$. 
For time-invariant Hawkes processes, we set the infectivity $\{\psi_{cc'}(t)\}$ to be $4$ constants randomly generated in the range $[0, 0.2]$. 
For time-varying Hawkes processes, we set $\psi_{cc'}(t)=0.2\cos(2\pi\frac{\omega_{cc'}}{50} t)$, where $\{\omega_{cc'}\}$ are randomly generated in the range $[1,4]$. 
Given these complete event sequences, we select $1000$ sequences as testing set while the remaining $1000$ sequences as training set. 
To generate SDC event sequences, we segment time window into $10$ intervals, and just randomly preserve the data in one interval for each training sequences. 
We test all methods in $10$ trials and compare them on the relative error between real parameters $\bm{\Theta}$ and their estimation results $\widehat{\bm{\Theta}}$, i.e., $\frac{\|\bm{\Theta}-\widehat{\bm{\Theta}}\|_2}{\|\bm{\Theta}\|_2}$, and the log-likelihood of testing sequences. 

\textbf{Time-invariant Hawkes Processes.} Fig.~\ref{Exp_HP} shows the comparisons on log-likelihood and relative error for various methods. 
In Fig.~\ref{Exp_HP_My} we can find that compared with the learning results based on complete event sequences, the results based on SDC event sequences degrade a lot (lower log-likelihood and higher relative error) because of the loss of information.  
Our data synthesis method improves the learning results consistently with the increase of training sequences, which outperforms its bootstrap-based competitor~\cite{politis1994stationary} as well. 
To demonstrate the universality of our method, besides our EM-based algorithm, we apply our method to the Least Squares (LS) algorithm~\cite{eichler2016graphical}. 
Fig.~\ref{Exp_HP_LS} shows that our method also improves the learning results of the LS algorithm in the case of SDC event sequences. 
Both the log-likelihood and the relative error obtained from the stitched sequences approach to the results learned from complete sequences. 

\begin{figure}[t]
\centering
\includegraphics[width=0.95\linewidth]{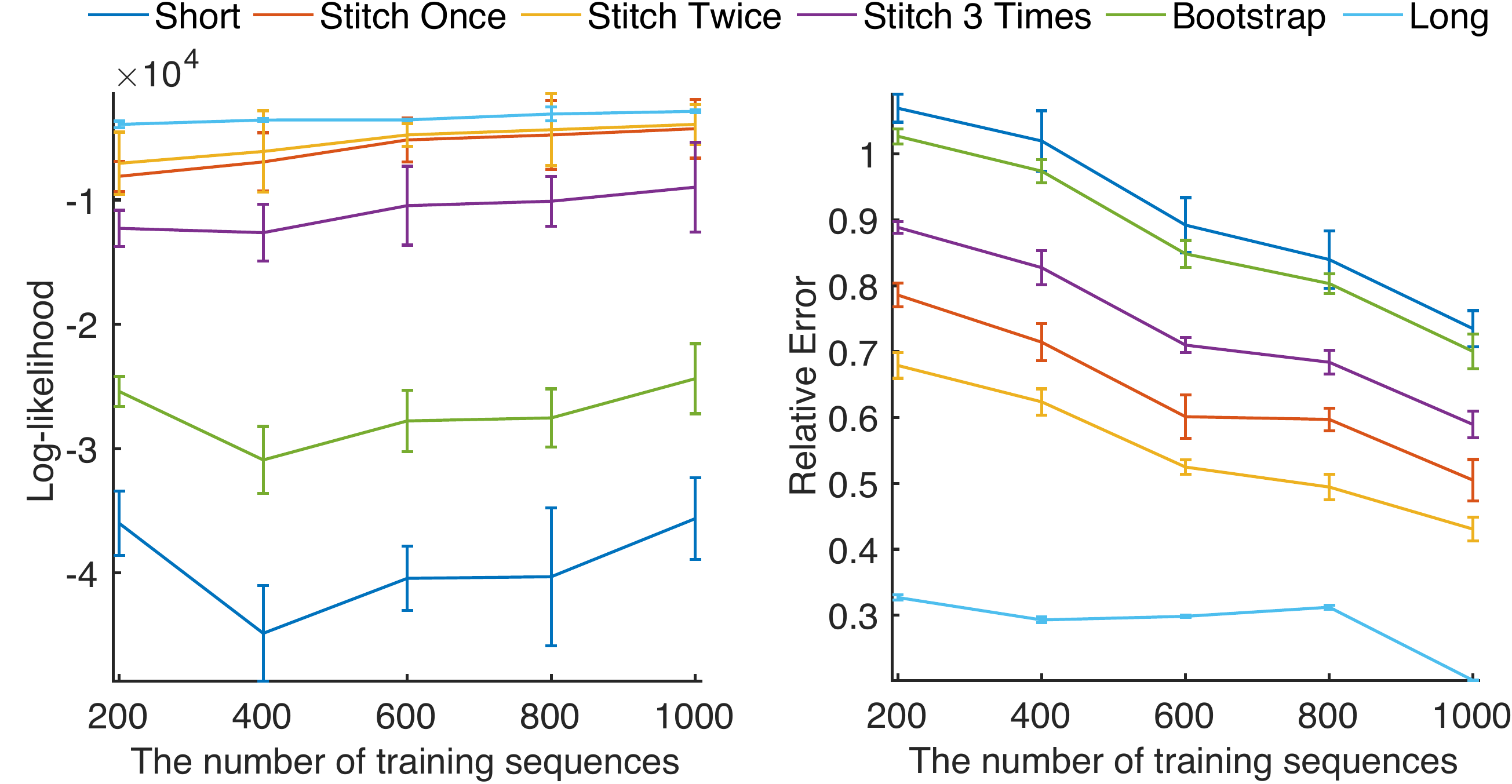}\vspace{-7pt}
\caption{Comparisons on log-likelihood and relative error.}\label{Exp1_TVHP}
\end{figure}
\begin{figure*}[t]
\centering
\includegraphics[width=1\linewidth]{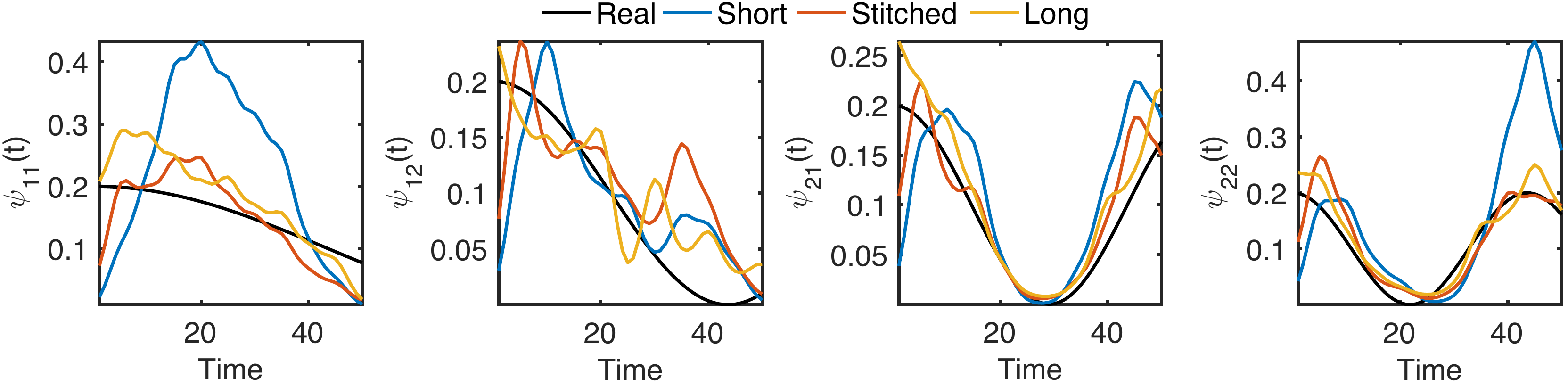}\vspace{-7pt}
\caption{Comparisons on infectivity functions $\{\psi_{cc'}(t)\}$. The number of original SDC sequences is $200$ and stitched via our method once.}\label{Exp2_TVHP}
\end{figure*}

\textbf{Time-varying Hawkes Processes.} 
Fig.~\ref{Exp1_TVHP} shows the comparisons on log-likelihood and relative error for various methods. 
Similarly, the learning results are improved because of applying our method --- higher log-likelihood and lower relative error are obtained and their standard deviation (the error bars associated with curves) is shrunk.
In this case, applying our method twice achieves better results than applying once, which verifies the usefulness of the iterative framework in our sampling-stitching algorithm. 
Besides objective measurements, in Fig.~\ref{Exp2_TVHP} we visualize the infectivity functions $\{\psi_{cc'}(t)\}$. 
It is easy to find that the infectivity functions learned from stitched sequences (red curves) are comparable to those learned from complete event sequences (yellow curves), which have small estimation errors of the ground truth (black curves). 

Note that our iterative framework is useful, especially for time-varying Hawkes processes, when the number of stitches is not very large. 
In our experiments, we fixed the maximum number of synthetic sequences. 
As a result, Figs.~\ref{Exp_HP} and \ref{Exp1_TVHP} show that the likelihoods first increase (i.e., stitching once or twice) and then decrease (i.e., stitching more than three times) while the relative errors have opponent changes  w.r.t. the number of stitches.
These phenomena imply that too many stitches introduce too much unreliable interdependency among events. 
Therefore, we fix the number of stitches to $2$ in the following experiments.

\begin{figure}[t]
\centering
\subfigure[LinkedIn data]{
\includegraphics[width=0.47\linewidth]{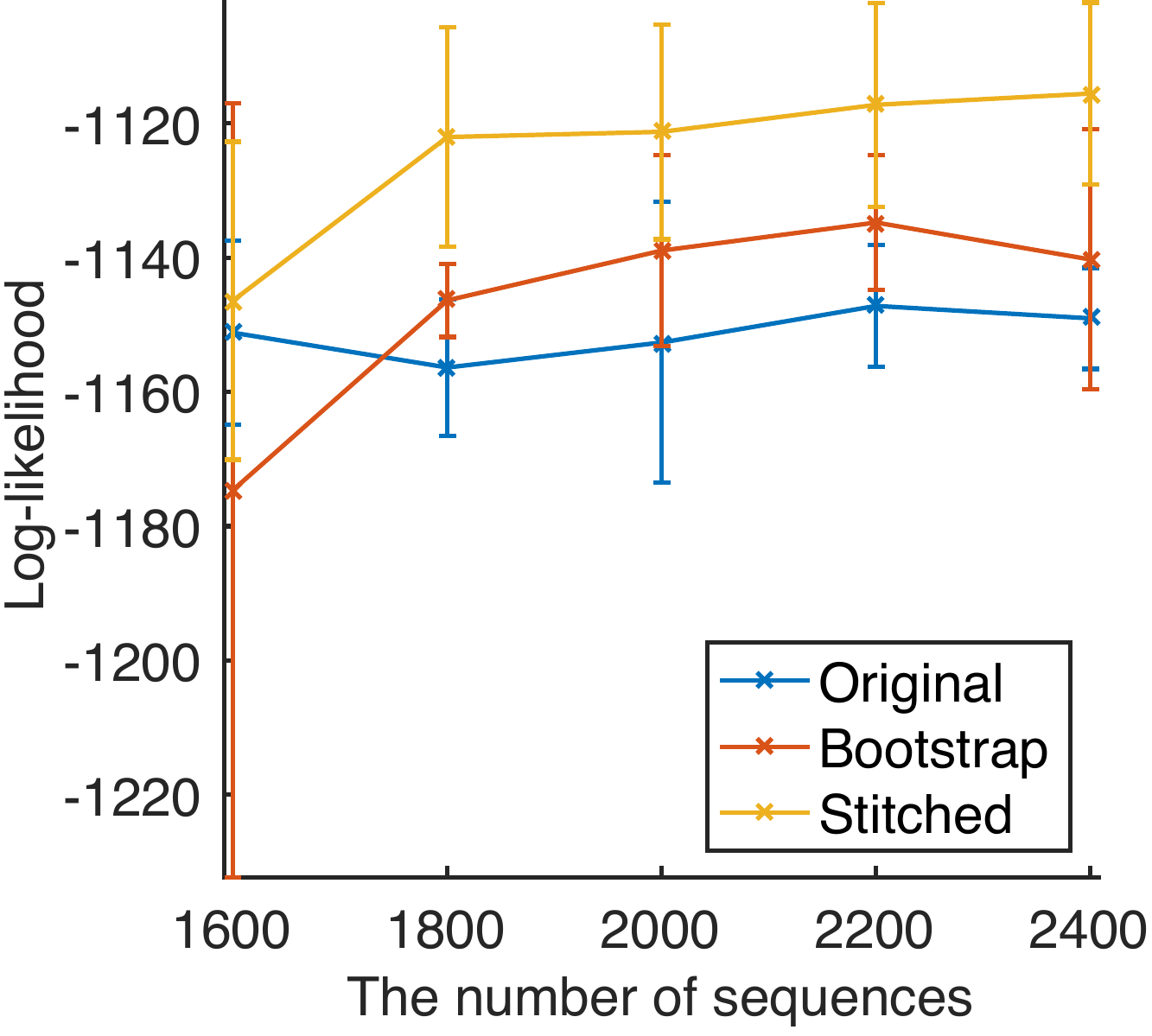}\label{link}
}
\subfigure[MIMIC III data]{
\includegraphics[width=0.47\linewidth]{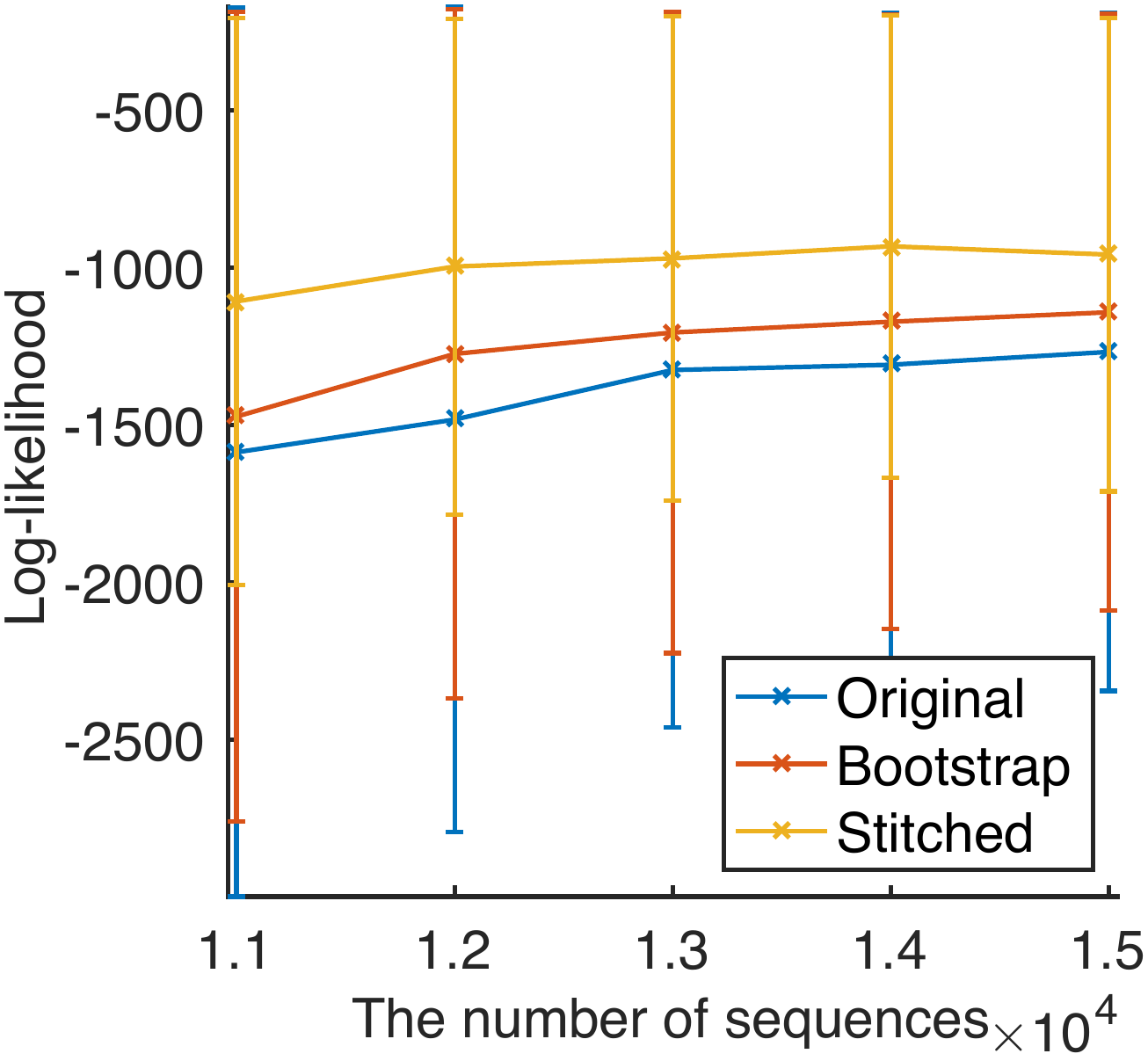}\label{mimic}
}
\vspace{-7pt}
\caption{Comparisons on log-likelihood.}
\end{figure}

\subsection{Real-World Data}\label{ssec:real}
Besides synthetic data, we also test our method on real-world data, including the LinkedIn data collected by ourselves and the MIMIC III data set~\cite{johnson2016mimic}.

\textbf{LinkedIn Data.} The LinkedIn data we collected online contain job hopping records of $3,000$ LinkedIn users in $82$ IT companies. 
For each user, her/his check-in time stamps corresponding to different companies are recorded as an event sequence, and her/his profile (e.g., education background, skill list, etc.) is treated as the feature associated with the sequence. 
For each person, the attractiveness of a company is always time-varying. 
For example, a young man may be willing to join in startup companies and increase his income via jumping between different companies. 
With the increase of age, he would more like to stay in the same company and achieve internal promotions. 
In other words, the infectivity network among different companies should be dynamical w.r.t. the age of employees. 
Unfortunately, most of the records in LinkedIn are short and doubly-censored --- only the job hopping events in recent years are recorded. 
How to construct the dynamical infectivity network among different companies from the SDC event sequences is still an open problem. 

\begin{figure*}[t]
\centering
\subfigure[LinkedIn data]{
\includegraphics[width=0.96\linewidth]{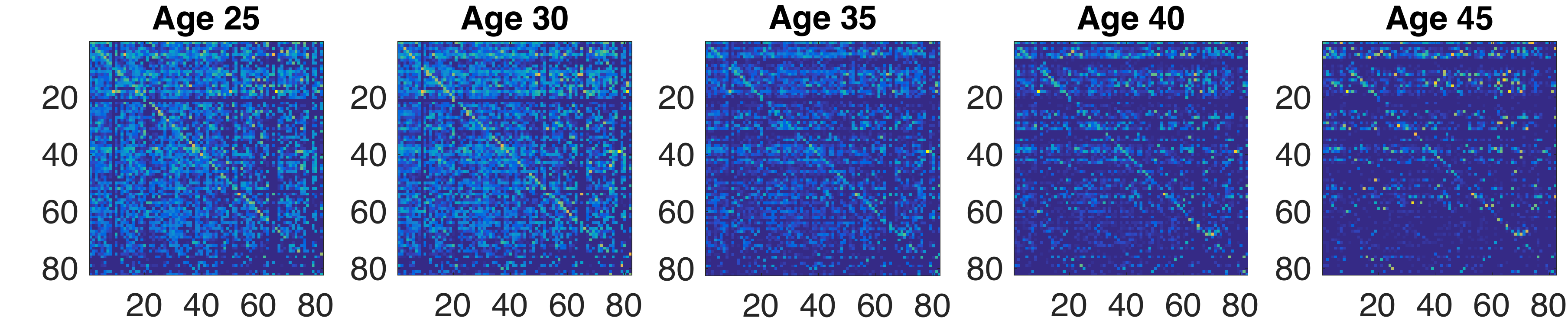}\label{link2}
}
\subfigure[MIMIC III data]{
\includegraphics[width=0.96\linewidth]{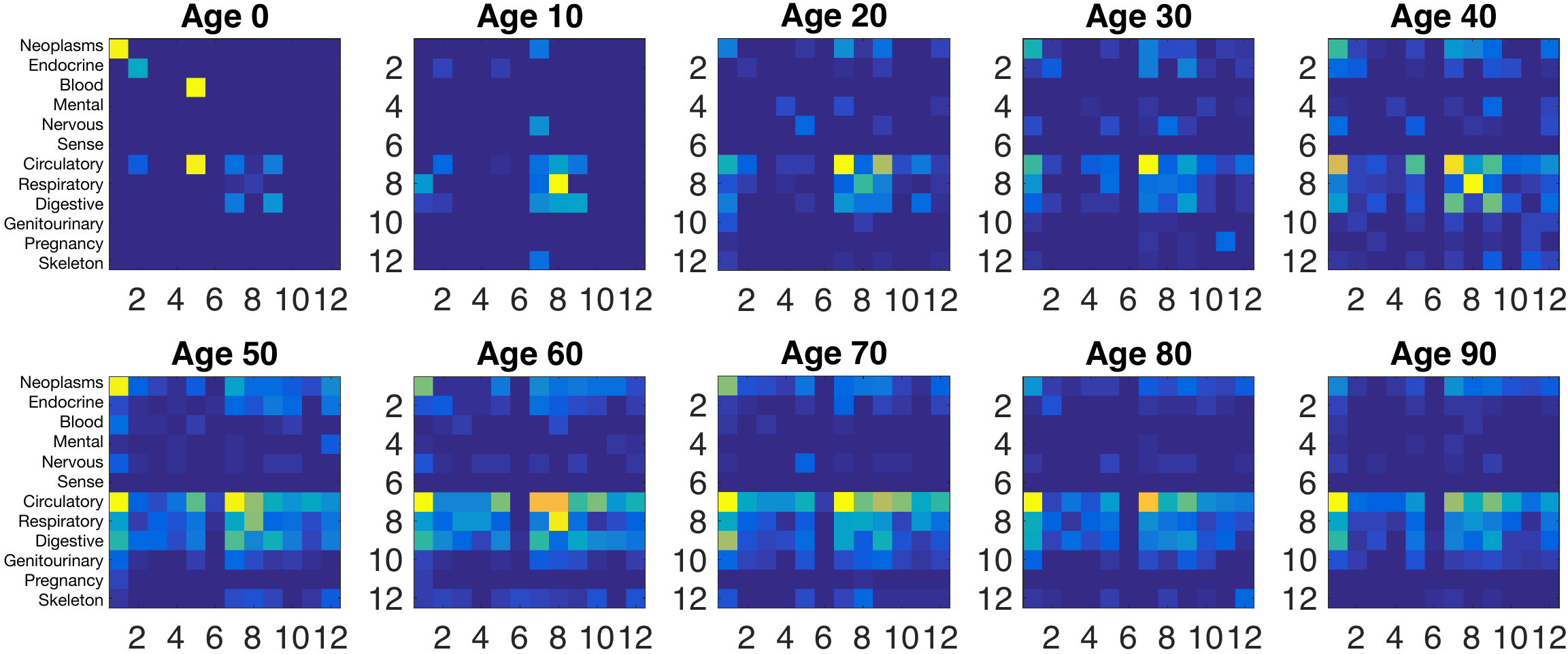}\label{mimic2}
}\vspace{-7pt}
\caption{Comparisons on infectivity functions $\{\psi_{cc'}(t)\}$.}
\end{figure*}

Applying our data synthesis method, we can stitch different users' job hopping sequences based on their ages (time stamps) and their profile (feature) and learn the dynamical network of company over time. 
In particular, we select $100$ users with relatively complete job hopping history (i.e., the range of their working experience is over $25$ years) as testing set. 
The remaining $2,900$ users are randomly selected as training set. 
The log-likelihood of testing set in $10$ trials is shown in Fig.~\ref{link}. 
We can find that the log-likelihood obtained from stitched sequences is higher than that obtained from original SDC sequences or that obtained from the sequences generated via the bootstrap method~\cite{politis1994stationary}, and its standard deviation is bounded stably. 
Fig.~\ref{link2} visualizes the adjacent matrix of infectivity network. 
The properties of the network verifies the rationality of our results: 1) the diagonal elements of the adjacent matrix are larger than other elements in general, which reflects the fact that most employees would like to stay in the same company and achieve a series of internal promotions; 2) with the increase of age, the infectivity network becomes sparse, which reflects the fact that users are more likely to try different companies in the early stages of their careers.

\textbf{MIMIC III Data.} The MIMIC III data contain admission records of over $40,000$ patients in the Beth Israel Deaconess Medical Center between 2001 and 2012. 
For each patient, her/his admission time stamps and diseases (represented via the ICD-9 codes~\cite{deyo1992adapting}) are recorded as an event sequence, and her/his profile (including gender, race and chronic history) is represented as binary feature of the sequence. 
As aforementioned, some work~\cite{choi2015constructing} has been done to extract time-invariant disease network from admission records. 
However, the real disease network should be time-varying w.r.t. the age of patient. 
Similar to the LinkedIn data, here we only obtain SDC event sequences --- the ranges of most admission records are just $1$ or $2$ years.

Applying our data synthesis method, we can leverage the information from different patients and stitch their sequences based on their ages and their profile. 
Focusing on $600$ common diseases in $12$ categories, we select $15,000$ patients' admission records randomly as training set and $1,000$ patients with relatively complete records as testing set. 
Fig.~\ref{mimic} shows that applying our data synthesis method indeed helps to improve log-likelihood of testing data. 
Compared with our bootstrap-based competitor, our data synthesis method gets more obvious improvements. 
Furthermore, we visualize the adjacent matrix of dynamical network of disease categories in Fig.~\ref{mimic2}. 
We can find that: 1) with the increase of age the disease network becomes dense, which reflects the fact that the complications of diseases are more and more common when people become old; 2) the networks show that neoplasms and the diseases of circulatory, respiratory, and digestive systems have strong self-triggering patterns because the treatments of these diseases often include several phases and require patients to make admission multiple times; 3) for kids and teenagers, their disease networks (i.e., ``Age 0'' and ``Age 10'' networks) are very sparse, and their common diseases mainly include neoplasms and the diseases of circulatory, respiratory, and digestive systems; 4) for middle-aged people, the reasons for their admissions are diverse and complicated so that their disease networks are dense and include many mutually-triggering patterns; 5) for longevity people, their disease networks (i.e., ``Age 80'' and ``Age 90'' networks) are relatively sparser than those of middle-aged people, because their admissions are generally caused by elderly chronic diseases. 

\begin{figure*}[t!]
\centering
\includegraphics[width=0.32\linewidth]{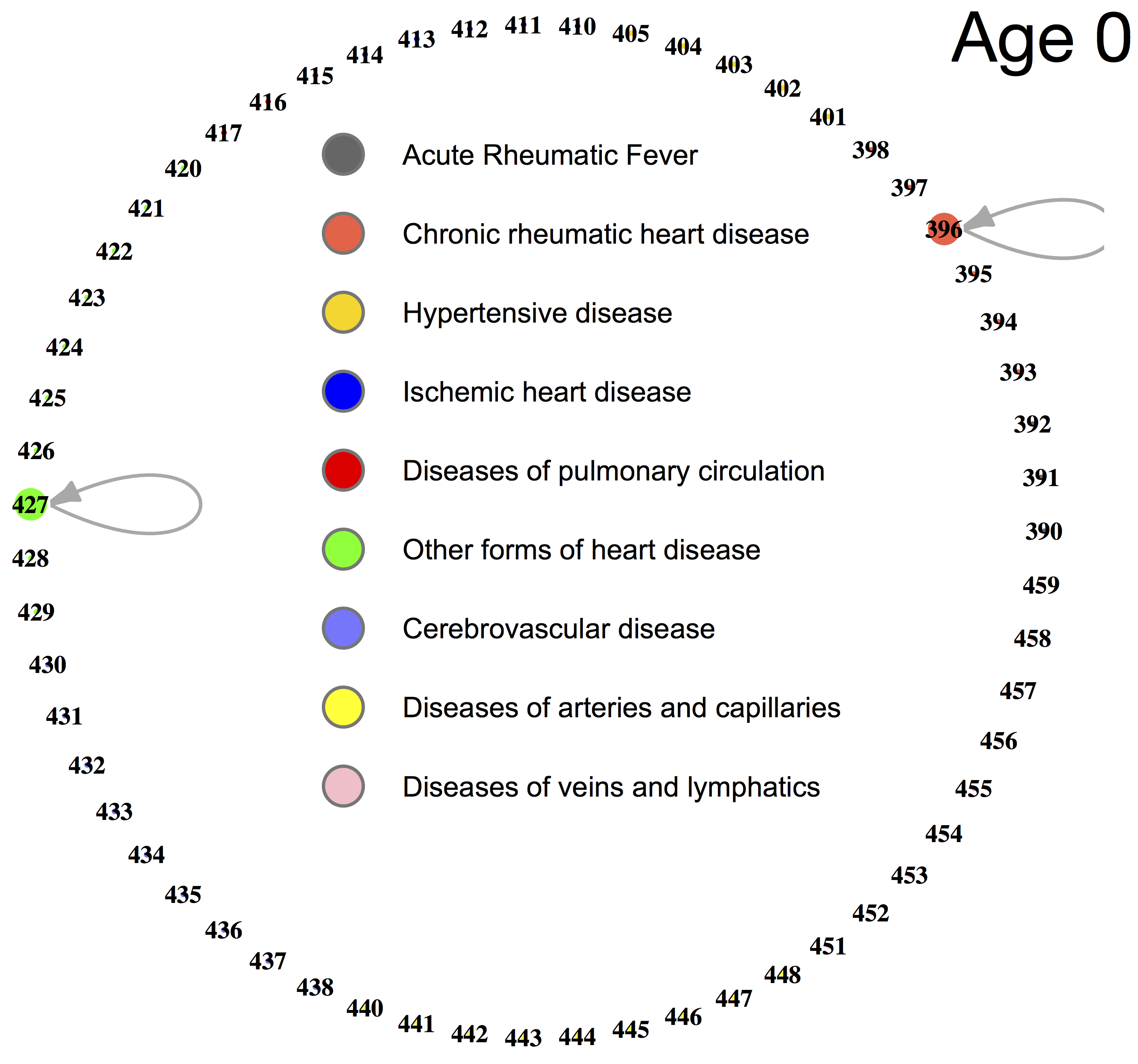}
\includegraphics[width=0.32\linewidth]{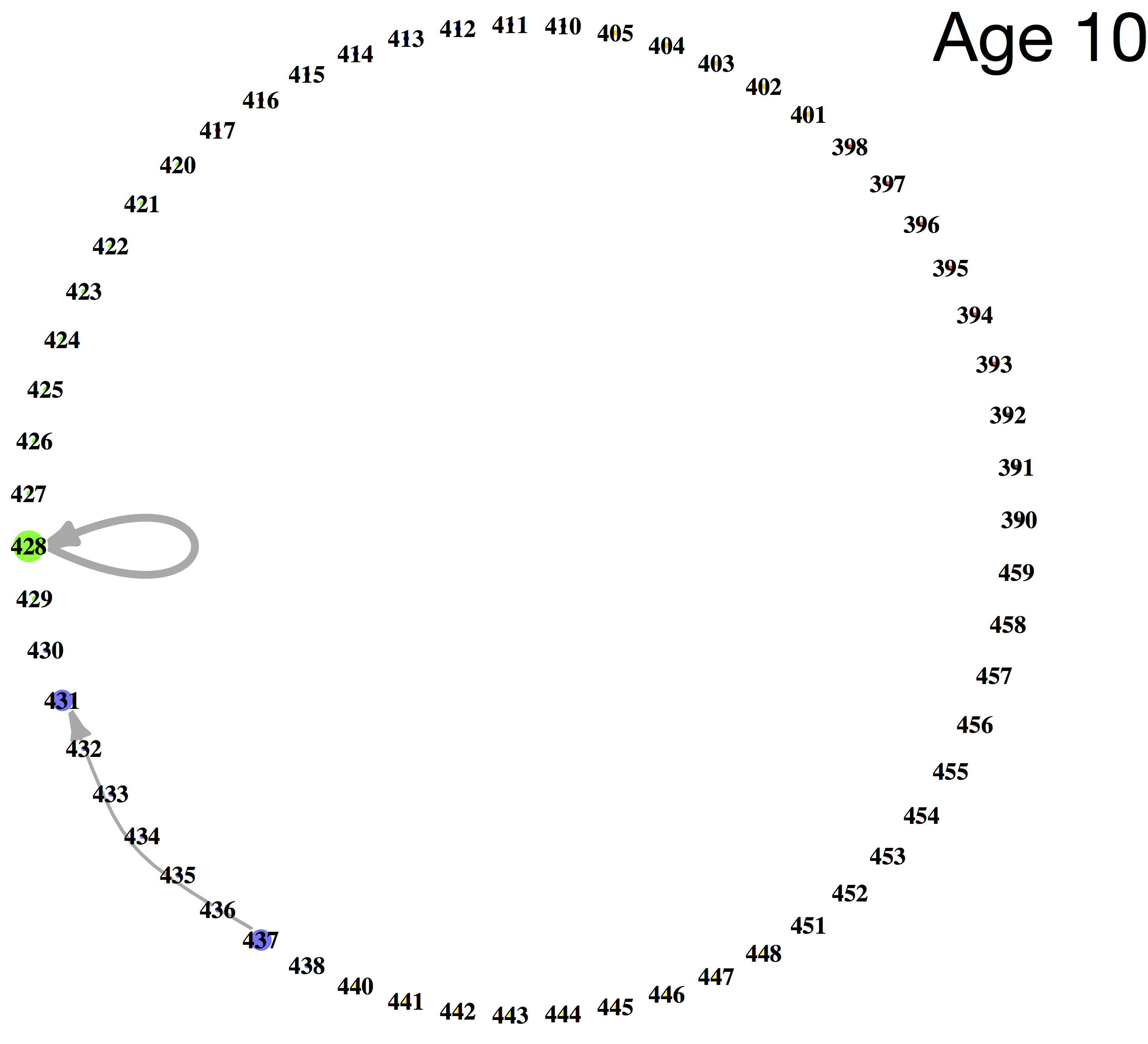}
\includegraphics[width=0.32\linewidth]{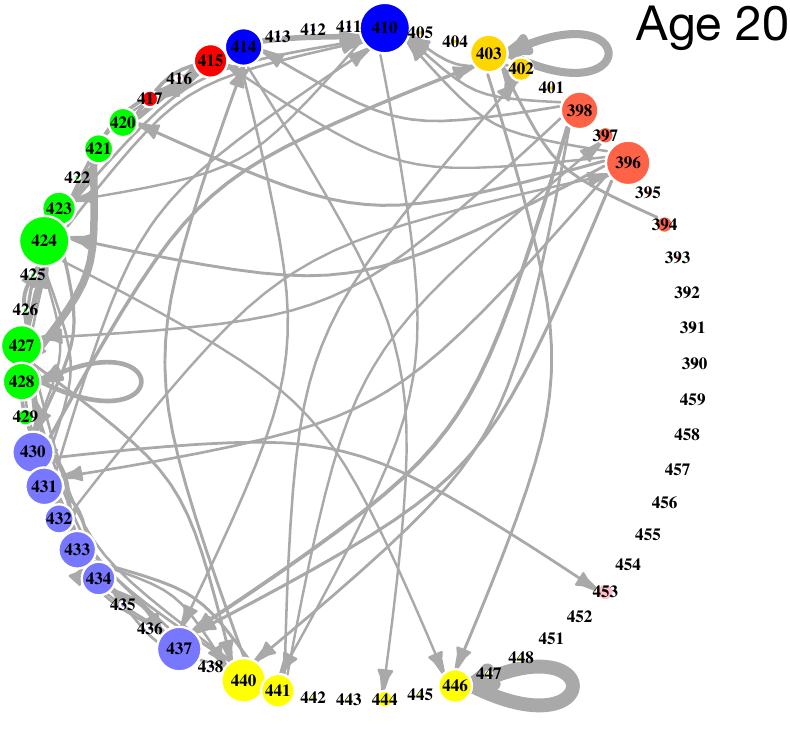}\\
\includegraphics[width=0.32\linewidth]{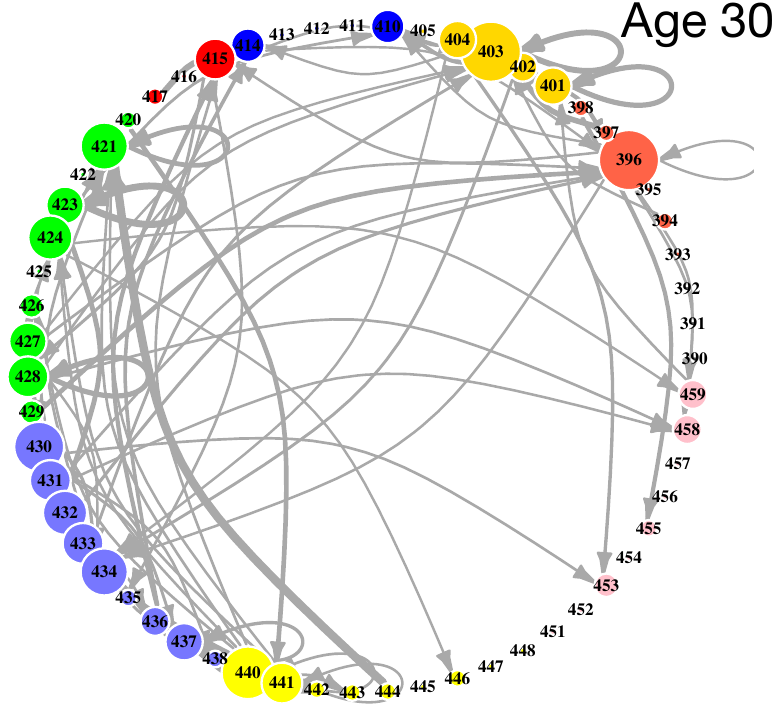}
\includegraphics[width=0.32\linewidth]{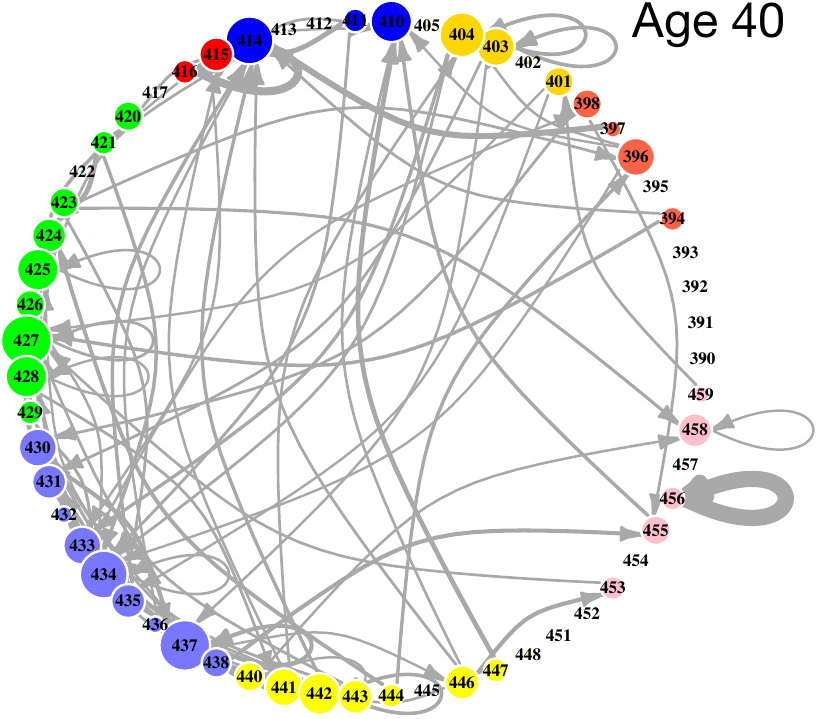}
\includegraphics[width=0.32\linewidth]{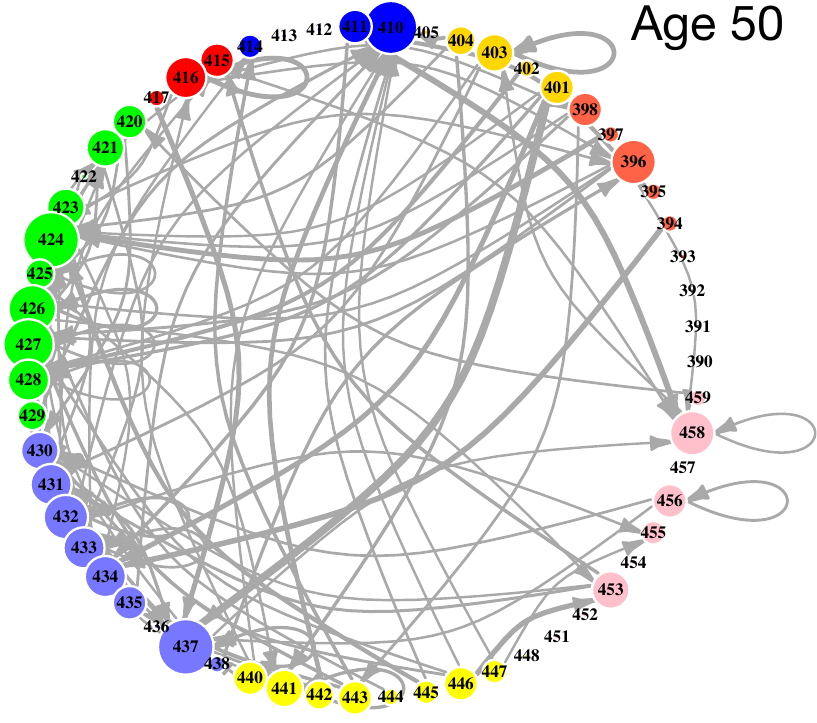}\\
\includegraphics[width=0.32\linewidth]{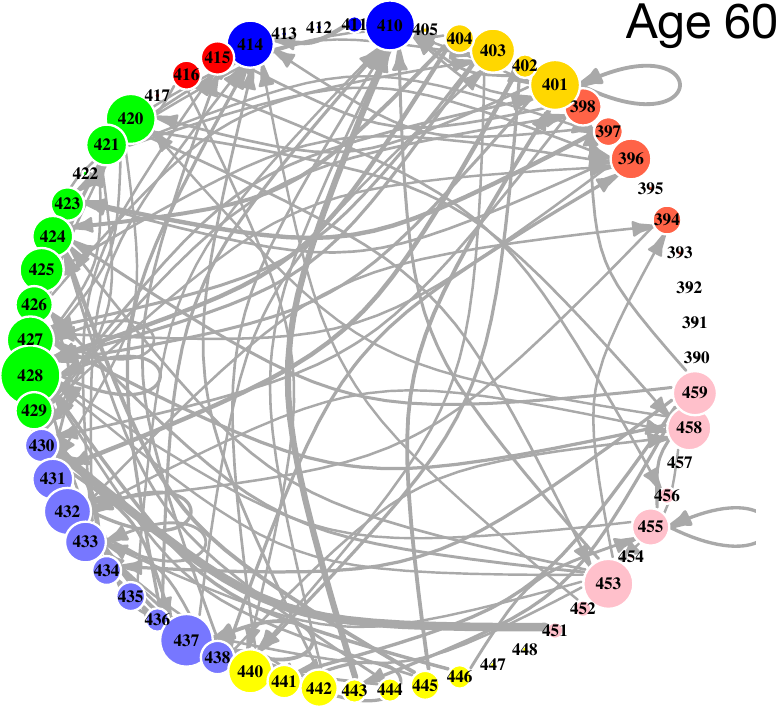}
\includegraphics[width=0.32\linewidth]{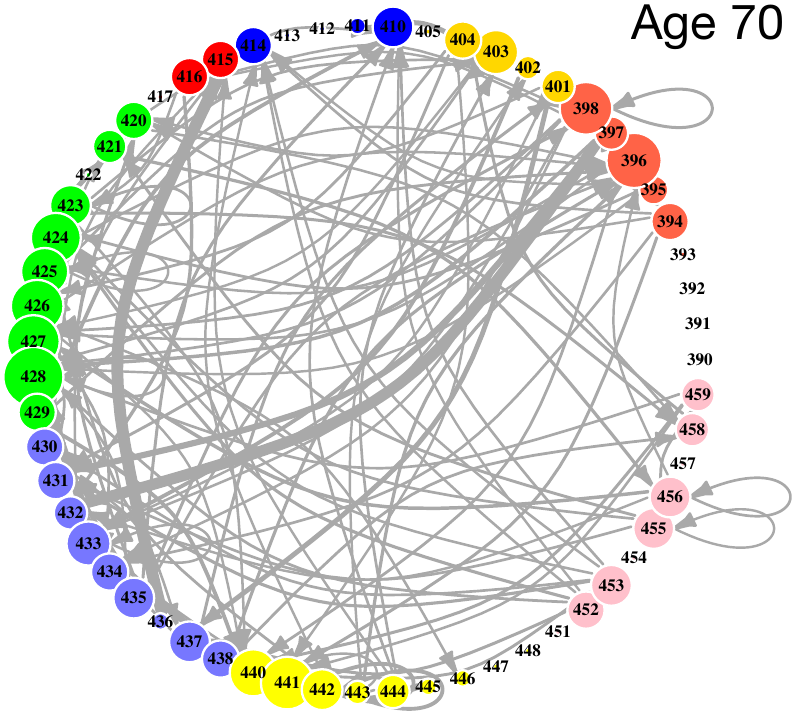}
\includegraphics[width=0.32\linewidth]{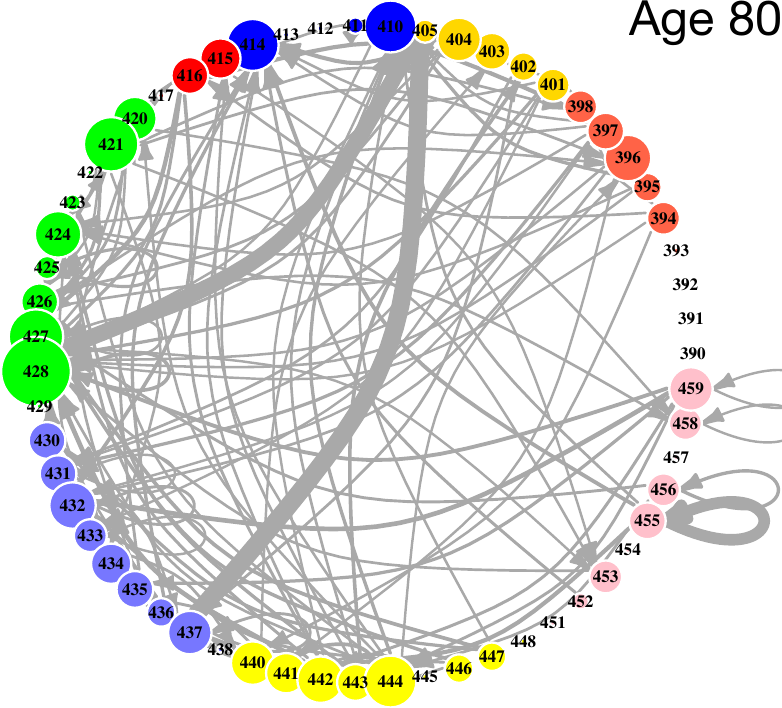}
\vspace{-7pt}
\caption{The diseases (nodes) are labeled with ICD-9 codes.  The colors indicate their sub-categories. 
The size of the $c$-th node is $\sum_{c'}\psi_{cc'}(t)$). 
The width of edge is $\psi_{cc'}(t)$.}\label{graph}
\end{figure*}

Additionally, we visualize the dynamical networks of the diseases of circulatory systems in~Fig.~\ref{graph}, and find some interesting triggering patterns. 
For example, for kids (``Age 0'' network), the typical circulatory diseases are ``diseases of mitral and aortic valves'' (ICD-9 396) and ``cardiac dysrhythmias'' (ICD-9 427), which are common for premature babies and the kids having congenital heart disease. 
For the old (``Age 80'' network), the network becomes dense. 
We can find that 1) as a main cause of death, ``heart failure'' (ICD-9 428) is triggered via multiple other diseases, especially ``secondary hypertension'' (ICD-9 405); 2) ``secondary hypertension'' is also likely to cause ``other and ill-defined cerebrovascular disease'' (ICD-9 437); 3) ``Hemorrhoids'' (ICD-9 455), as a common disease with strong self-triggering pattern, will cause frequent admissions of patients. 
In summary, the analysis above verifies the rationality of our result --- the dynamical disease networks we learned indeed reflect the properties of human's health trajectory. 

\section{Conclusion}
\label{sec:con}
In this paper, we propose a novel data synthesis method to learn Hawkes processes from SDC event sequences. 
With the help of temporal information and optional features, we measure the similarities among different SDC event sequences and estimate the distribution of potential long event sequences. 
Applying a sampling-stitching mechanism, we successfully synthesize a large amount of long event sequences and learn point processes robustly. 
We test our method for both time-invariant and time-varying Hawkes processes. 
Experiments show that our data synthesis method improves the robustness of learning algorithms for various models. 
In the future, we plan to provide more theoretical and quantitative analysis to our data synthesis method and apply it to more applications.

\textbf{Acknowledgements.} 
This work is supported in part via NSF DMS-1317424, NIH R01 GM108341, NSFC 61628203, U1609220 and the Key Program of Shanghai Science and Technology Commission 15JC1401700.

\bibliography{example_paper}

\begin{thebibliography}{35}
\providecommand{\natexlab}[1]{#1}
\providecommand{\url}[1]{\texttt{#1}}
\expandafter\ifx\csname urlstyle\endcsname\relax
  \providecommand{\doi}[1]{doi: #1}\else
  \providecommand{\doi}{doi: \begingroup \urlstyle{rm}\Url}\fi

\bibitem[Bacry et~al.(2013)Bacry, Delattre, Hoffmann, and Muzy]{bacry2013some}
Bacry, Emmanuel, Delattre, Sylvain, Hoffmann, Marc, and Muzy, Jean-Francois.
\newblock Some limit theorems for hawkes processes and application to financial
  statistics.
\newblock \emph{Stochastic Processes and their Applications}, 123\penalty0
  (7):\penalty0 2475--2499, 2013.

\bibitem[Choi et~al.(2015)Choi, Du, Chen, Song, and Sun]{choi2015constructing}
Choi, Edward, Du, Nan, Chen, Robert, Song, Le, and Sun, Jimeng.
\newblock Constructing disease network and temporal progression model via
  context-sensitive hawkes process.
\newblock In \emph{ICDM}, 2015.

\bibitem[Cowling et~al.(1996)Cowling, Hall, and Phillips]{cowling1996bootstrap}
Cowling, Ann, Hall, Peter, and Phillips, Michael~J.
\newblock Bootstrap confidence regions for the intensity of a poisson point
  process.
\newblock \emph{Journal of the American Statistical Association}, 91\penalty0
  (436):\penalty0 1516--1524, 1996.

\bibitem[Daley \& Vere-Jones(2007)Daley and Vere-Jones]{daley2007introduction}
Daley, Daryl~J and Vere-Jones, David.
\newblock \emph{An introduction to the theory of point processes: volume II:
  general theory and structure}.
\newblock Springer Science \& Business Media, 2007.

\bibitem[De~Gruttola \& Lagakos(1989)De~Gruttola and Lagakos]{de1989analysis}
De~Gruttola, Victor and Lagakos, Stephen~W.
\newblock Analysis of doubly-censored survival data, with application to aids.
\newblock \emph{Biometrics}, pp.\  1--11, 1989.

\bibitem[Deyo et~al.(1992)Deyo, Cherkin, and Ciol]{deyo1992adapting}
Deyo, Richard~A, Cherkin, Daniel~C, and Ciol, Marcia~A.
\newblock Adapting a clinical comorbidity index for use with icd-9-cm
  administrative databases.
\newblock \emph{Journal of clinical epidemiology}, 45\penalty0 (6):\penalty0
  613--619, 1992.

\bibitem[Efron(1982)]{efron1982jackknife}
Efron, Bradley.
\newblock \emph{The jackknife, the bootstrap and other resampling plans},
  volume~38.
\newblock SIAM, 1982.

\bibitem[Eichler et~al.(2016)Eichler, Dahlhaus, and
  Dueck]{eichler2016graphical}
Eichler, Michael, Dahlhaus, Rainer, and Dueck, Johannes.
\newblock Graphical modeling for multivariate hawkes processes with
  nonparametric link functions.
\newblock \emph{Journal of Time Series Analysis}, 2016.

\bibitem[Fan(2009)]{fan2009local}
Fan, Chun-Po~Steve.
\newblock \emph{Local Likelihood for Interval-censored and Aggregated Point
  Process Data}.
\newblock PhD thesis, University of Toronto, 2009.

\bibitem[Gon{\c{c}}alves \& Kilian(2004)Gon{\c{c}}alves and
  Kilian]{gonccalves2004bootstrapping}
Gon{\c{c}}alves, S{\i}́lvia and Kilian, Lutz.
\newblock Bootstrapping autoregressions with conditional heteroskedasticity of
  unknown form.
\newblock \emph{Journal of Econometrics}, 123\penalty0 (1):\penalty0 89--120,
  2004.

\bibitem[Guan \& Loh(2007)Guan and Loh]{guan2007thinned}
Guan, Yongtao and Loh, Ji~Meng.
\newblock A thinned block bootstrap variance estimation procedure for
  inhomogeneous spatial point patterns.
\newblock \emph{Journal of the American Statistical Association}, 102\penalty0
  (480):\penalty0 1377--1386, 2007.

\bibitem[Hawkes \& Oakes(1974)Hawkes and Oakes]{hawkes1974cluster}
Hawkes, Alan~G and Oakes, David.
\newblock A cluster process representation of a self-exciting process.
\newblock \emph{Journal of Applied Probability}, pp.\  493--503, 1974.

\bibitem[Johnson et~al.(2016)Johnson, Pollard, Shen, Lehman, Feng, Ghassemi,
  Moody, Szolovits, Celi, and Mark]{johnson2016mimic}
Johnson, Alistair~EW, Pollard, Tom~J, Shen, Lu, Lehman, Li-wei~H, Feng,
  Mengling, Ghassemi, Mohammad, Moody, Benjamin, Szolovits, Peter, Celi,
  Leo~Anthony, and Mark, Roger~G.
\newblock Mimic-iii, a freely accessible critical care database.
\newblock \emph{Scientific data}, 3, 2016.

\bibitem[Kirk \& Stumpf(2009)Kirk and Stumpf]{kirk2009gaussian}
Kirk, Paul~DW and Stumpf, Michael~PH.
\newblock Gaussian process regression bootstrapping: exploring the effects of
  uncertainty in time course data.
\newblock \emph{Bioinformatics}, 25\penalty0 (10):\penalty0 1300--1306, 2009.

\bibitem[Klein \& Moeschberger(2005)Klein and Moeschberger]{klein2005survival}
Klein, John~P and Moeschberger, Melvin~L.
\newblock \emph{Survival analysis: techniques for censored and truncated data}.
\newblock Springer Science \& Business Media, 2005.

\bibitem[Kobayashi \& Lambiotte(2016)Kobayashi and
  Lambiotte]{kobayashi2016tideh}
Kobayashi, Ryota and Lambiotte, Renaud.
\newblock Tideh: Time-dependent hawkes process for predicting retweet dynamics.
\newblock \emph{arXiv preprint arXiv:1603.09449}, 2016.

\bibitem[Luo et~al.(2015)Luo, Xu, Zhen, Ning, Zha, Yang, and
  Zhang]{luo2015multi}
Luo, Dixin, Xu, Hongteng, Zhen, Yi, Ning, Xia, Zha, Hongyuan, Yang, Xiaokang,
  and Zhang, Wenjun.
\newblock Multi-task multi-dimensional hawkes processes for modeling event
  sequences.
\newblock In \emph{IJCAI}, 2015.

\bibitem[Luo et~al.(2016)Luo, Xu, Zhen, Dilkina, Zha, Yang, and
  Zhang]{luo2016learning}
Luo, Dixin, Xu, Hongteng, Zhen, Yi, Dilkina, Bistra, Zha, Hongyuan, Yang,
  Xiaokang, and Zhang, Wenjun.
\newblock Learning mixtures of markov chains from aggregate data with
  structural constraints.
\newblock \emph{Transactions on Knowledge and Data Engineering}, 28\penalty0
  (6):\penalty0 1518--1531, 2016.

\bibitem[Mei \& Eisner(2016)Mei and Eisner]{mei2016neural}
Mei, Hongyuan and Eisner, Jason.
\newblock The neural hawkes process: A neurally self-modulating multivariate
  point process.
\newblock \emph{arXiv preprint arXiv:1612.09328}, 2016.

\bibitem[Paparoditis \& Politis(2001)Paparoditis and
  Politis]{paparoditis2001tapered}
Paparoditis, Efstathios and Politis, Dimitris~N.
\newblock Tapered block bootstrap.
\newblock \emph{Biometrika}, 88\penalty0 (4):\penalty0 1105--1119, 2001.

\bibitem[Politis \& Romano(1994)Politis and Romano]{politis1994stationary}
Politis, Dimitris~N and Romano, Joseph~P.
\newblock The stationary bootstrap.
\newblock \emph{Journal of the American Statistical association}, 89\penalty0
  (428):\penalty0 1303--1313, 1994.

\bibitem[Rubin(2009)]{rubin2009multiple}
Rubin, Donald~B.
\newblock \emph{Multiple Imputation for Nonresponse in Surveys}, volume 307.
\newblock John Wiley \& Sons, 2009.

\bibitem[Streit(2010)]{streit2010poisson}
Streit, Roy~L.
\newblock \emph{Poisson point processes: imaging, tracking, and sensing}.
\newblock Springer Science \& Business Media, 2010.

\bibitem[Sun \& Kalbfleisch(1995)Sun and Kalbfleisch]{sun1995estimation}
Sun, J and Kalbfleisch, JD.
\newblock Estimation of the mean function of point processes based on panel
  count data.
\newblock \emph{Statistica Sinica}, pp.\  279--289, 1995.

\bibitem[Turnbull(1974)]{turnbull1974nonparametric}
Turnbull, Bruce~W.
\newblock Nonparametric estimation of a survivorship function with doubly
  censored data.
\newblock \emph{Journal of the American Statistical Association}, 69\penalty0
  (345):\penalty0 169--173, 1974.

\bibitem[Van~den Berg \& Drepper(2016)Van~den Berg and
  Drepper]{van2016inference}
Van~den Berg, Gerard~J and Drepper, Bettina.
\newblock Inference for shared-frailty survival models with left-truncated
  data.
\newblock \emph{Econometric Reviews}, 35\penalty0 (6):\penalty0 1075--1098,
  2016.

\bibitem[Wellner \& Zhang(2000)Wellner and Zhang]{wellner2000two}
Wellner, Jon~A and Zhang, Ying.
\newblock Two estimators of the mean of a counting process with panel count
  data.
\newblock \emph{Annals of Statistics}, pp.\  779--814, 2000.

\bibitem[Xu et~al.(2015)Xu, Zhen, and Zha]{xu2015trailer}
Xu, Hongteng, Zhen, Yi, and Zha, Hongyuan.
\newblock Trailer generation via a point process-based visual attractiveness
  model.
\newblock In \emph{IJCAI}, 2015.

\bibitem[Xu et~al.(2016{\natexlab{a}})Xu, Farajtabar, and Zha]{xu2016learning}
Xu, Hongteng, Farajtabar, Mehrdad, and Zha, Hongyuan.
\newblock Learning granger causality for hawkes processes.
\newblock In \emph{ICML}, 2016{\natexlab{a}}.

\bibitem[Xu et~al.(2016{\natexlab{b}})Xu, Ning, Zhang, Rhee, and
  Jiang]{xu2016pinfer}
Xu, Hongteng, Ning, Xia, Zhang, Hui, Rhee, Junghwan, and Jiang, Guofei.
\newblock Pinfer: Learning to infer concurrent request paths from system kernel
  events.
\newblock In \emph{ICAC}, 2016{\natexlab{b}}.

\bibitem[Xu et~al.(2016{\natexlab{c}})Xu, Wu, Nemati, and Zha]{xu2016icu}
Xu, Hongteng, Wu, Weichang, Nemati, Shamim, and Zha, Hongyuan.
\newblock Icu patient flow prediction via discriminative learning of
  mutually-correcting processes.
\newblock \emph{arXiv preprint arXiv:1602.05112}, 2016{\natexlab{c}}.

\bibitem[Yang \& Zha(2013)Yang and Zha]{yang2013mixture}
Yang, Shuang-Hong and Zha, Hongyuan.
\newblock Mixture of mutually exciting processes for viral diffusion.
\newblock In \emph{ICML}, 2013.

\bibitem[Zhao et~al.(2015)Zhao, Erdogdu, He, Rajaraman, and
  Leskovec]{zhao2015seismic}
Zhao, Qingyuan, Erdogdu, Murat~A, He, Hera~Y, Rajaraman, Anand, and Leskovec,
  Jure.
\newblock Seismic: A self-exciting point process model for predicting tweet
  popularity.
\newblock In \emph{KDD}, 2015.

\bibitem[Zhou et~al.(2013{\natexlab{a}})Zhou, Zha, and Song]{zhou2013learning}
Zhou, Ke, Zha, Hongyuan, and Song, Le.
\newblock Learning social infectivity in sparse low-rank networks using
  multi-dimensional hawkes processes.
\newblock In \emph{AISTATS}, 2013{\natexlab{a}}.

\bibitem[Zhou et~al.(2013{\natexlab{b}})Zhou, Zha, and Song]{zhou2013learning2}
Zhou, Ke, Zha, Hongyuan, and Song, Le.
\newblock Learning triggering kernels for multi-dimensional hawkes processes.
\newblock In \emph{ICML}, 2013{\natexlab{b}}.

\end{thebibliography}
\bibliographystyle{icml2016}

\end{document}